\documentclass[12pt,a4paper]{article}
\pdfoutput=0  
\usepackage[margin=32mm]{geometry}
\usepackage{graphicx}
\usepackage{psfrag}
\usepackage{natbib}
\usepackage{url}
\usepackage{authblk}
\usepackage[utf8]{inputenc}
\usepackage{amsfonts,amsmath,amssymb,bm}
\usepackage{dsfont}
\usepackage{mathrsfs}
\usepackage{mathtools}
\usepackage{multirow,enumerate}
\usepackage[dvipsnames]{xcolor}
\usepackage{tikz}
\usepackage{tocloft}  
\usepackage[nottoc]{tocbibind}

\usepackage{hyperref,bookmark}
\usepackage[english]{babel}
\usepackage[autostyle, english = american]{csquotes}
\MakeOuterQuote{"}

\newcommand \tit {Bayesian sequential design of computer experiments for quantile set inversion}

\newcommand \kwA {Gaussian processes}
\newcommand \kwB {Active learning}
\newcommand \kwC {Design of computer experiments}
\newcommand \kwD {Stepwise Uncertainty Reduction}
\newcommand \kwE {Set inversion}
\newcommand \kwF {Uncertainty quantification}
\newcommand \kw  {\kwA, \kwB, \kwC, \kwD, \kwE, \kwF}

\hypersetup{
  bookmarksnumbered,
  colorlinks=true,
  allcolors=Violet,
  pdfauthor={Romain Ait Abdelmalek-Lomenech, Julien Bect, Vincent Chabridon, Emmanuel Vazquez},
  pdftitle={\tit},
  pdfsubject={},
  pdfkeywords={\kw}}

\newtheorem{remark}{Remark}

\newcommand{\ProbExpectLetter}[1]{\mathsf{#1}}

\renewcommand{\P}{\ProbExpectLetter{P}}
\newcommand{\Ps}{\P_{S}}
\newcommand{\PtildeS}{\P_{\tilde S}}
\newcommand{\Pn}{\P_n}

\newcommand{\E}{\ProbExpectLetter{E}}
\newcommand{\En}{\E_n}

\newcommand{\X}{\mathds{X}}
\renewcommand{\S}{\mathds{S}}
\newcommand{\Uset}{\mathds{U}}
\newcommand{\Rset}{\mathds{R}}
\newcommand{\Gf}{\Gamma(f)}

\newcommand{\Gxi}{\Gamma(\xi)}

\newcommand{\Hn}{\mathcal{H}_n}
\newcommand{\Hnn}{\mathcal{H}_{n+1}}

\newcommand{\In}{\mathcal{I}_n}
\newcommand{\mn}{\mu_n}
\newcommand{\sn}{\sigma_n}
\newcommand{\hGn}{\widehat\Gamma_n}

\newcommand{\tS}{\widetilde\S}
\newcommand{\tX}{\widetilde\X}
\newcommand{\tD}{\widetilde{\mathds{U}}}
\newcommand{\dx}{\mathrm{d}x}
\newcommand{\ds}{\mathrm{d}s}

\newcommand{\one}{\mathds{1}}
\newcommand{\Zobs}{Z^{\mathrm{obs}}}
\newcommand{\Xc}{\breve x}
\newcommand{\Sc}{\breve s}

\DeclareMathOperator{\card}{card}
\DeclareMathOperator*{\argmin}{argmin}
\DeclareMathOperator*{\argmax}{argmax}

\psfrag{steps}[c][c]{\raisebox{-3pt}{steps}}
\psfrag{prop}[c][c]{\raisebox{10pt}{prop. misclass.}}
\psfrag{random}{\small random}
\psfrag{Ranjan}{\small Ranjan}
\psfrag{misclassification}{\small misclass.}
\psfrag{ECL}{\small ECL}
\psfrag{Jointsur}{\small Joint-SUR}
\psfrag{Qsisur}{\small QSI-SUR}

\newcommand{\ticksize}{\scriptsize}
\psfrag{-0.02} {\ticksize $-0.02$}
\psfrag{0}     {\ticksize $0$}
\psfrag{0.02}  {\ticksize $0.02$}
\psfrag{0.025} {\ticksize $0.025$}
\psfrag{0.04}  {\ticksize $0.04$}
\psfrag{0.05}  {\ticksize $0.05$}
\psfrag{0.06}  {\ticksize $0.06$}
\psfrag{0.075} {\ticksize $0.075$}
\psfrag{0.08}  {\ticksize $0.08$}
\psfrag{0.1}   {\ticksize $0.1$}
\psfrag{0.12}  {\ticksize $0.12$}
\psfrag{0.14}  {\ticksize $0.14$}
\psfrag{0.16}  {\ticksize $0.16$}
\psfrag{0.125} {\ticksize $0.125$}
\psfrag{0.15}  {\ticksize $0.15$}
\psfrag{0.2}   {\ticksize $0.2$}
\psfrag{0.25}  {\ticksize $0.25$}
\psfrag{0.3}   {\ticksize $0.3$}
\psfrag{0.35}  {\ticksize $0.35$}
\psfrag{0.4}   {\ticksize $0.4$}
\psfrag{0.5}   {\ticksize $0.5$}
\psfrag{0.6}   {\ticksize $0.6$}
\psfrag{0.7}   {\ticksize $0.7$}
\psfrag{0.8}   {\ticksize $0.8$}
\psfrag{0.9}   {\ticksize $0.9$}
\psfrag{1}     {\ticksize $1$}
\psfrag{5}     {\ticksize $5$}
\psfrag{10}    {\ticksize $10$}
\psfrag{15}    {\ticksize $15$}
\psfrag{20}    {\ticksize $20$}
\psfrag{25}    {\ticksize $25$}
\psfrag{30}    {\ticksize $30$}
\psfrag{40}    {\ticksize $40$}
\psfrag{50}    {\ticksize $50$}
\psfrag{60}    {\ticksize $60$}
\psfrag{80}    {\ticksize $80$}
\psfrag{100}   {\ticksize $100$}
\psfrag{150}   {\ticksize $150$}
\psfrag{200}   {\ticksize $200$}
\psfrag{250}   {\ticksize $250$}
\psfrag{300}   {\ticksize $300$}

\psfrag{bigbigmiscbased}{\small misc.-based}
\psfrag{bigbigvarbased}{\small var.-based}
\psfrag{bigbigentrbased}{\small entr.-based}

\psfrag{X}{$x$}
\psfrag{S}{$s$}

\newlength{\toto}

\usetikzlibrary{patterns}
\usepackage{pgfplots}
\newcommand{\setA}{plot [smooth cycle, tension=1] coordinates {(0,0) (2,1.5) (3,0) (2,-1)}}
\newcommand{\setB}{plot [smooth cycle, tension=1] coordinates {(1.5,-1) (3.5,1.5) (4.5,0) (3.5,-1)}}

\cftsetindents{section}{1em}{2.8em}
\cftsetindents{subsection}{3.8em}{3.1em}
\cftsetindents{subsubsection}{6.9em}{4.7em}

\begin{document}

\pdfbookmark[section]{Title page}{titlepage}

\def\spacingset#1{\renewcommand{\baselinestretch}%
{#1}\small\normalsize} \spacingset{1}


  \title{\bf \tit} %
  \author[1]{Romain Ait Abdelmalek-Lomenech\thanks{%
      Corresponding author.

      \medbreak

      \noindent The authors
      gratefully acknowledge the National French Research Agency (ANR)
      for funding this work in the context of the SAMOURAI project
      (ANR-20-CE46-0013). The authors report there are no competing interests to declare.

      \medbreak
      
      \noindent This is an author-generated postprint version.\\
      \noindent Accepted for publication in Technometrics, %
      \href{https://doi.org/10.1080/00401706.2024.2394475}{DOI:10.1080/00401706.2024.2394475}%
    }} %
  \author[1]{Julien Bect}
  \author[2]{Vincent~Chabridon}
  \author[1]{Emmanuel Vazquez}

  \affil[1]{\small Université Paris-Saclay, CNRS, CentraleSupélec,

    Laboratoire des Signaux et Systèmes, 91190 Gif-sur-Yvette, France\raisebox{-1pt}{$\vphantom{\big|}$}}
  \affil[2]{\small EDF R\&D, 6 Quai Watier, 78401 Chatou, France}

  \date{}

  \maketitle

\vspace{-3mm}
\noindent\rule{\textwidth}{3pt}

\begin{abstract}
  We consider an unknown multivariate function representing a
  system---such as a complex numerical simulator---taking both
  deterministic and uncertain inputs. %
  Our objective is to estimate the set of deterministic inputs leading
  to outputs whose probability (with respect to the distribution of
  the uncertain inputs) of belonging to a given set is less than a
  given threshold. %
  This problem, which we call Quantile Set Inversion (QSI), occurs for
  instance in the context of robust (reliability-based)
  optimization problems, when looking for the set of solutions that
  satisfy the constraints with sufficiently large probability. %
  To solve the QSI problem we propose a Bayesian strategy, based on
  Gaussian process modeling and the Stepwise Uncertainty Reduction
  (SUR) principle, to sequentially choose the points at which the
  function should be evaluated to efficiently approximate the set of
  interest. %
  We illustrate the performance and interest of the proposed SUR
  strategy through several numerical experiments.
\end{abstract}

\noindent\rule{\textwidth}{3pt}

\vspace{15mm}

\noindent%
{\it Keywords:} \kw. \vspace*{5mm}

\cleardoublepage
\pdfbookmark[section]{Table of contents}{toc}
\tableofcontents
 
\cleardoublepage
\spacingset{1.25}

\section{Introduction}
\label{sec:intro}

When dealing with a numerical model of a physical phenomenon or a
system, one is often interested in estimating the set of input
parameters leading to outputs in a given range. %
Such \emph{set inversion} problems \citep{jaulin:1993:automatica}
arise in various frameworks. %
In particular, "robust" formulations of the set inversion problem, in
which some inputs are considered uncertain, have appeared recently in
the literature, with applications to nuclear safety
\citep{chevalier:2013:phdthesis, marrel:2022:icscream}, %
flood defense optimization \citep{richet:2019:inversion} %
and pollution control systems \citep{elamri:2023:set-inversion}.

Following \cite{richet:2019:inversion}, we focus on a robust
formulation of the set inversion problem that we call \emph{quantile
  set inversion} (QSI). %
We consider a system modeled by an unknown continuous function
$f \, : \, \X\times\S \to \Rset^q$, where $\X$ and $\S$ are bounded
subsets of $\Rset^{d_\X}$ and $\Rset^{d_\S}$, corresponding to the
sets of admissible values for the deterministic and uncertain (or
stochastic) input variables of the system. %
We model the uncertain inputs by a random vector~$S$ with known
distribution $\Ps$ on~$\S$.  Then, given a subset $C\subset\Rset^q$ of
the output space and a threshold $\alpha \in (0,1)$, our objective is
to estimate the set
\begin{equation}
  \label{eq:gammaf}
  \Gf = \{x\in\X\,:\,\P(f(x,S) \in C) \le \alpha\}.
\end{equation}
Using the language of machine learning, we can also formulate the QSI
problem as that of learning a classifier~$\X \to \{ 0, 1 \}$ as close
as possible to the indicator function~$\one_{\Gf}$. %
The QSI problem occurs for instance in the context of robust
(reliability-based, a.k.a. chance-constrained) optimization problems,
when looking for the set of solutions that violate the constraints
with sufficiently small probability---where, with our notations, the
constraints are violated when~$f(x, S)$ belongs to the critical
region~$C$.

\begin{figure}[h]
  \centering
  \psfrag{0}{\tiny $0$}
  \psfrag{0.1}{\tiny $0.1$}
  \psfrag{0.2}{\tiny $0.2$}
  \psfrag{0.3}{\tiny $0.3$}
  \psfrag{0.4}{\tiny $0.4$}
  \psfrag{0.5}{\tiny $0.5$}
  \psfrag{0.6}{\tiny $0.6$}
  \psfrag{0.8}{\tiny $0.8$}
  \psfrag{1}{\tiny $1$}
  \psfrag{2}{\tiny $2$}
  \psfrag{-2}{\tiny $-2$}
  \psfrag{4}{\tiny $4$}
  \psfrag{6}{\tiny $6$}
  \psfrag{8}{\tiny $8$}
  \psfrag{5}{\tiny $5$}
  \psfrag{10}{\tiny $10$}
  \psfrag{12}{\tiny $12$}
  \psfrag{14}{\tiny $14$}
  \psfrag{15}{\tiny $15$}
  \psfrag{16}{\tiny $16$}
  \psfrag{18}{\tiny $18$}
  \psfrag{density}{\footnotesize \raisebox{-4pt}{density}}
  \psfrag{  boundaryboundaryboundary}{\footnotesize boundary of $f^{-1}(C)$}
  \psfrag{  Gamma}{\footnotesize $\Gamma(f)$}
  \psfrag{  indicator}{\footnotesize $\mathds{1}_{\Gf} (x)$}
  \psfrag{  proba}{\footnotesize $\P(f(x,S) \in C)$}
  \psfrag{  alpha}{\footnotesize $\alpha$}
  \includegraphics[width=\textwidth,height=6.5cm]{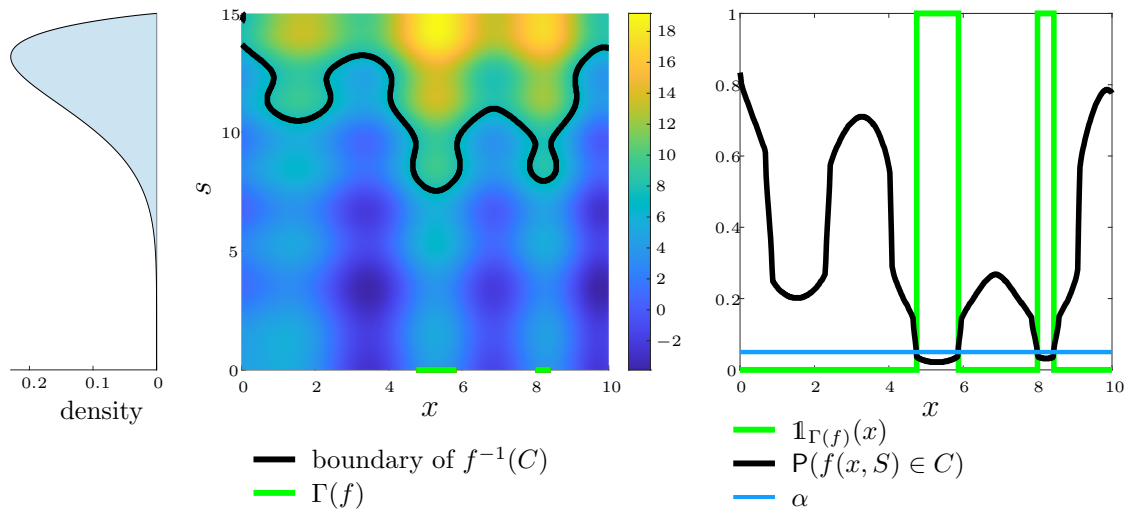}
  \caption{%
    Representation of a two-dimensional QSI problem. %
    Left: probability density function of~$\Ps$. %
    Middle: test function~$f = f_1$ (see \autoref{sec:presentation}
    for details), boundary of~$f^{-1}(C)$ and quantile set~$\Gamma(f)$
    associated to $C = (-\infty, 7.5]$ and $\alpha = 0.05$. %
    Right: indicator function of the quantile set,
    probability~$\P(f(x,S) \in C)$ and probability
    threshold~$\alpha$.}
  \label{fig:f1}
\end{figure}

An illustrative two-dimensional example of a QSI problem is shown
in~\autoref{fig:f1}, %
with one deterministic input variable and one uncertain input variable
($d_\X = d_\S = 1$), %
critical region $C = (-\infty, 7.5]$ in the output space, %
and probability threshold $\alpha = 5\%$. %
The input space $\S = \left[ 0, 15 \right]$ for the uncertain variable
is equipped with a $\mathrm{Beta}(7.5, 1.9)$ distribution, rescaled
from $[0,1]$ to~$\S$, which concentrates on large values of~$S$. %
The set~$\Gamma(f)$ to be estimated is the union of two disjoint
intervals in~$\X = \left[ 0, 10 \right]$.  %
It appears clearly, on this example, that an accurate approximation of
the boundary of~$f^{-1}(C)$, in $\X \times \S$, is only needed in some
specific regions of the input space---more specifically, for the
points~$(x, s)$ of the boundary such that the probability
$\P(f(x,S) \in C)$ is close to the threshold~$\alpha$. %
(See \autoref{sec:presentation} for numerical results on this
example.)

\begin{remark}
  When $q=1$ and $C$ is a semi-infinite interval, there is a direct
  link between $\Gf$ and the quantiles of $f$. For instance, if
  $C= (-\infty, T]$, the set~\eqref{eq:gammaf} can be rewritten as
  \begin{equation}
    \label{eq:Gamma-rewritten-quantile}
    \Gamma(f) = \{x \in \X \, : \, Q_{\alpha}(f(x,S)) > T \},
  \end{equation}
  where $Q_{\alpha}(f(x,S))$ denotes the quantile of order $\alpha$ of
  $f(x,S)$, with $S\sim\Ps$. %
  More generally, $\Gamma(f)$ can be seen as a quantile of the random
  set~$\{x \in \X \, : \, f(x,S) \in C \}$ in the sense of
  \cite{molchanov:1991:quantiles}---hence our choice of terminology.
\end{remark}

When the numerical model $f$ is computationally expensive, it is
important to estimate~$\Gf$ using only a small number of evaluations
of~$f$. %
With this constraint in mind, we propose in this article a sequential
Bayesian strategy based on the \emph{Stepwise Uncertainty Reduction}
(SUR) principle \citep[see, e.g.,][]{vazquez2009sequential,
  villemonteix:2009:informational, bect:2012:sur_failureproba,
  chevalier:2014:fast_parallel_kriging}. %
The starting point of a SUR strategy is to view~$f$ as a sample path
of a random process, in practice a Gaussian process (GP). %
Then, at each step, an evaluation point is chosen by minimizing
  the \emph{expected future uncertainty} on the quantity or object of
  interest---a set in the present case---given the past observations.

The structure of the article is as follows: %
\autoref{sec:bayesian} introduces the framework, while
\autoref{sec:overview} gives a brief overview of the literature on Bayesian set
inversion strategies, with a particular emphasis on SUR approaches. %
The core contribution of the article is given in
\autoref{sec:construction}, which presents the construction of a SUR
sampling criterion for the QSI problem. %
\autoref{sec:numerical} demonstrates the performance of our approach
on various numerical examples, including an application to
history matching. In \autoref{sec:conclusion}, we summarize our
conclusions and provide perspectives for further research.

\medbreak

\itshape
\paragraph{Nota bene. }%
The authors have become aware, at the occasion of the SIAM Conference
on Uncertainty Quantification (UQ22) in Atlanta, of related research
work conducted by Charlie~Sire (IRSN, France) and co-authors
\citep{sire:2022:siam}. %
The research presented in this article has been carried out
independently of theirs.  \normalshape

\section{Framework and notations}
\label{sec:bayesian}


In the following, we consider a function $f : \Uset \to \Rset$, where
$\Uset = \X$ or $\Uset = \X\times\S$, depending on whether there are
stochastic input variables or not. %
We adopt a Bayesian approach to sequentially choose the evaluation
points $U_1, U_2,\,\ldots\in\Uset$ of $f$ and estimate $\Gf$ from
evaluation results. %
It is assumed that we observe, at each selected point~$U_n$, a
response $\Zobs_n = f(U_n) + \epsilon_n$, where the $\epsilon_n$ are
independent zero-mean Gaussian random variables, with a
possibly null variance in the case of a deterministic
simulator. %
As a prior for the unknown function~$f$, we consider a GP~model
\citep[see, e.g.,][]{rasmussen:2006:gpml, santner:2018:design}---in
other words, we assume that $f$~is a sample path of a~GP. %
We denote by $\xi$ this process, and by $\mu$ and $k$ its mean and
covariance functions.%

Denote by~$\In = \{(U_1, \Zobs_1),\, \ldots,\, (U_n, \Zobs_n)\}$ the
currently available information, and $\Pn = \P(\, \cdot \, | \, \In)$
the conditional probability given $\In$. %
Bayesian strategies employ at each step a \emph{sampling
  criterion}, also referred to as an \emph{acquisition function},
which we will denote by~$J_n$ when it is meant to be minimized,
or~$G_n$ when it is meant to be maximized. %
This criterion, based on the distribution of~$\xi$ under~$\Pn$, is
used to select the next evaluation point from~$\Uset$. %
More explicitly: we choose $U_{n+1}$ as an element in~$\Uset$ that
minimizes~$J_n$ or maximizes~$G_n$:
\begin{equation*}
  U_{n+1} \in \argmin_{u \in \Uset} J_n(u) \quad \text{or} \quad
  U_{n+1} \in \argmax_{u \in \Uset} G_n(u).
\end{equation*}
In the following sections, two families of such criteria are reviewed.

\paragraph{Notations. } %
In the rest of the paper, $\En = \E\left(\cdot \mid \In\right)$
denotes the conditional expectation associated with~$\P_n$, %
$\mn(u)$ and $\sn(u)$ stand for the conditional (posterior) mean and
standard deviation of~$\xi(u)$, %
and $p_n(u) = \Pn(\xi(u)\in C)$ is the conditional (posterior)
probability that $\xi(u)$~belongs to~$C$.

\section{Overview of Bayesian strategies for set inversion}
\label{sec:overview}

\subsection{Maximal uncertainty sampling}
\label{sec:pointwise}

We review in this section a first family of sampling criteria, which
corresponds to the general idea of \emph{maximal uncertainty
  sampling}, i.e., sampling at the location~$x \in \X$ where the
uncertainty about~$\one_C(\xi(x))$ and/or~$\xi(x)$ is maximal. %
The literature on such criteria only deals, to the best of our
knowledge, with the deterministic case $\Uset = \X$, when $f$ is a
real-valued function ($q=1$), and when $C=(T,+\infty)$, for a given
$T\in \Rset$. %
In this setting, the set inversion problem reduces to the estimation
of the set
\begin{equation}
  \Lambda(f) = \{x \in \X \, : \, f(x) \le T \}.
\end{equation}
A natural approach to this problem is to select the point at which the
probability of misclassification is maximal \citep{bryan:2005:active},
leading to the sampling criterion
$G_n(x) = \min\left(p_n(x), 1-p_n(x)\right)$. %
This criterion is maximal for any point~$x$ such that~$\mu_n(x) =
T$. %
Several equivalent criteria lead to the same choice of sampling point,
including the entropy of the indicator~$\mathds{1}_C(\xi(x))$ used
by~\cite{cole:2023:entropy}, its variance, %
or the sampling criterion used in the AK-MCS method
of~\cite{echard:AK-MCS:2011}.

Other sampling criteria operate a trade-off between the posterior
variance of~$\xi(x)$ and its estimated proximity to the
threshold~$T$. %
This is the case, for instance, for the family of criteria defined by
$G_n(x) = \En\left[\max\left(0, (\kappa\sn(x))^\delta-
    \left|\xi(x)-T\right|^\delta\right)\right]$, with $\kappa > 0$,
introduced separately by \cite{bichon:2008} with~$\delta = 1$, and
\cite{ranjan:2008:contour} with~$\delta = 2$. %
Similarly, \cite{bryan:2005:active} proposed the \emph{straddle
  heuristic}, where $G_n(x) = 1.96\,\sn(x) - |\mn(x) - T|$.

\subsection{Stepwise uncertainty reduction}
\label{sec:sur}
SUR strategies \citep[see][and references
therein]{bect:2019:supermartingale} are a special case of the Bayesian
approach in which the evaluation points are sequentially chosen by
minimizing the \emph{expected future uncertainty} about the object of
interest. %
More precisely, a SUR strategy starts by defining a measure of
uncertainty $\Hn$, at each step, that depends on the currently
available information $\In$. %
Then, a sampling criterion $J_n$ is built by considering the
expectation of $\Hnn$ conditional on~$\mathcal{I}_{n}$, for a given
choice of~$U_{n+1} = u$:
\begin{equation}
  J_n(u) \;=\; \En \left[ \Hnn\mid  U_{n+1}= u \right].
\end{equation}
Notice that $\Hnn$ depends on the unknown outcome of the evaluation
at~$u$ and that $J_{n}(u)$ is an expectation over this random outcome.
Equivalently, instead of minimizing the sampling criterion~$J_n$, one
can maximize the information gain $G_{n}(u) = \Hn - J_n(u)$.%

We now give more details and first focus on the case of deterministic
inversion ($\Uset = \X$). %
Several approaches have been developed in the past years. %
For instance, \cite{bect:2012:sur_failureproba} suggest the integrated
probability of misclassification
\begin{equation}\label{equ:Hn:misclass}
  \Hn = \int_\X \min(p_n(x), 1-p_n(x))\, \dx,
\end{equation}
and $\Hn = \int_\X p_n(x)(1-p_n(x))\, \dx$, the integrated variance of
$\mathds{1}_{\Lambda(\xi)}(x)$ as uncertainty measures. %
Similarly, \cite{marques:2018:contour-entropy} propose to use the
integrated entropy of the random variable
$\mathds{1}_{(-\infty;T-\epsilon(x)]}(\xi(x))-
\mathds{1}_{[T+\epsilon(x);+\infty)}(\xi(x))$, with $\epsilon(x) > 0$,
to estimate $\Lambda(f)$ in a context where different sources of
information can be leveraged.

\cite{picheny:2010:adaptive_design} propose a \emph{targeted
  Integrated Mean Square Error} (tIMSE) reduction strategy,
based on the uncertainty measure:
$\Hn = \int_{\X}\sn^2(x)\, W_n(x)\, \dx$, %
where $W_n(x) = \En[K(\mn(x)-T)]$, with $K$ a kernel
(e.g., Gaussian or uniform).

For additional examples of uncertainty measures and corresponding SUR
criteria applicable to the deterministic set inversion problem, refer
to \cite{chevalier:2013:estimating_vorobev},
\cite{chevalier:2013:phdthesis}, \cite{azzimonti:2021:adaptive}, and
\cite{duhamel:2023:sur-bichon}.

To conclude this section, let us mention two formulations of the set
inversion problem with uncertain input variables, which are
related to---but distinct from---the QSI problem. %
First, \cite{chevalier:2013:phdthesis} considers the task of
estimating the set
$\{x \in \X \, : \, \underset{s\in\S}{\max}f(x,s) \le T\}$. %
In this setting, the proposed uncertainty measure is
$\Hn = \int_\X p^{\circ}_{n}(x)(1- p^{\circ}_n(x))\dx$, where
$p^{\circ}_n(x) = \Pn(\max_{s\in\S}\xi(x,s) \le T)$. %
Second, in the work of \cite{elamri:2023:set-inversion}, the objective
instead is to estimate the set $\{x\in\X \,:\, \E(f(x,S)) \le T\}$. %
To this end, the authors propose a hybrid SUR strategy to choose,
sequentially, the deterministic component~$x$ and the stochastic
component~$s$ of each new evaluation point.

\section{Construction of a SUR strategy for QSI}
\label{sec:construction}

\subsection{Sampling criterion}
\label{sec:design}

Our objective is now to estimate the set~$\Gamma(f)$ defined
by~\eqref{eq:gammaf} using evaluation results modeled by
$Z_n^{obs} = f(X_n, S_n) + \epsilon_n$, %
i.e., we take $\Uset = \X \times \S$ and write
$U_n = \left( X_n, S_n \right)$ from now on. %

In the following, we construct a SUR sampling criterion for the QSI
problem. %
For the sake of simplicity, we assume that the distribution~$\Ps$
admits a density $g$ (with respect to the Lebesgue measure). %
Consider the random process
\begin{equation}
  \tau(x) \;=\; \int_\S \one_C\left( \xi(x, s) \right)g(s)\, \ds,
\end{equation}
which corresponds, for each $x \in \X$, to the stochastic (Bayesian)
counterpart of the unknown probability~$\P\left( f(x, S) \in C \right)$,
and notice that $\Gxi$ can be written as
\begin{equation}
  \label{eq:Gf-with-tau}
  \Gxi = \left\{ x\in\X \, : \, \tau(x) \le \alpha \right\}\,.
\end{equation}

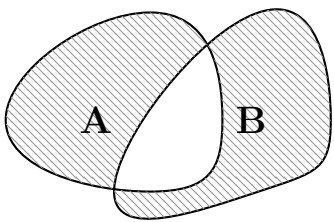
\begin{figure}
  \centering
  \begin{tikzpicture}[scale=0.95]
    \draw[thick] \setA;
    \draw[thick] \setB;
    \begin{scope}
      \fill[pattern=north west lines, pattern color = black!40, even odd rule] \setA \setB;
    \end{scope}
    \node at (1.25,0) {\large$\mathbf{A}$};
    \node at (3.4,0) {\large$\mathbf{B}$};
  \end{tikzpicture}
  \caption{%
    Symmetric difference
    $A\Delta B = (A \setminus B) \cup (B \setminus A)$ between $A$
    and~$B$ (shaded area).}
  \label{fig:symdiff}
\end{figure}

Assume that a sequence $(\hGn)_{n\geq 1}$ of estimators of $\Gxi$ has
been chosen. %
We propose to use as uncertainty measure the expected volume of the
symmetric difference (see \autoref{fig:symdiff}) between~$\Gxi$ and
its estimator:
\begin{equation}
  \label{eq:Hn1}
  \Hn = \En\left[\lambda(\Gxi\,\Delta\,\hGn)\right],
\end{equation}
where $\lambda$ is the usual (Lebesgue) volume measure
on~$\Rset^{d_\X}$.
The SUR strategy derived from~\eqref{eq:Hn1} consists in minimizing,
at each step, the criterion
\begin{equation}
  J_n(x, s) = \En\left[\Hnn \; | \; (X_{n+1},S_{n+1}) = (x, s)\right]\,.
\end{equation}
SUR strategies using the symmetric difference have been used, in other
contexts, by \cite{chevalier:2013:phdthesis}
and~\cite{azzimonti:2021:adaptive}.

When considering the Bayes-optimal estimator
\begin{equation}
  \label{eq:Bayes-opt-estim}
  \hGn = \left\{x\in\X \, : \, \pi_n(x)>\frac{1}{2}\right\},
\end{equation}
where $\pi_n(x) = \Pn(\tau(x) \le \alpha)$, %
the uncertainty measure $\Hn$ can be expressed as the integrated
probability of misclassification associated to the
classifier~$\mathds{1}_{\hGn}$:
\begin{equation}
  \label{eq:Bayes-opt-measure}
  \Hn = \int_\X\min(\pi_n(x),\, 1-\pi_n(x))\, \dx\,.
\end{equation}
The proof of this simple result can be found in 
(\autoref{SM:sec:proof-Hn}). %
As a consequence, the SUR strategy derived from~\eqref{eq:Hn1}
consists in minimizing at each step the criterion
\begin{equation}
  \label{eq:QSI-SUR-criterion}
  \begin{split}
    J_n(\Xc,\Sc)
    & = \En\left[ \Hnn \mid  (X_{n+1},S_{n+1}) = (\Xc,\Sc)\right]\,\\
    & = \int_\X\En\left[\min(\pi_{n+1}(x),\, 1-\pi_{n+1}(x)) \mid  (X_{n+1},S_{n+1}) = (\Xc,\Sc)\right]\, \dx\, .
  \end{split}
\end{equation}
Here, notice that we use the new notations~$\Xc$ and~$\Sc$ for the
components of the candidate point, since $x$~is now used as an
integration variable.

\begin{remark}\label{rem:other-uncertainty}
  Other choices for the uncertainty measure are possible. %
  Notably, we could use any increasing transformation of
  $\min(\pi_n(x),\, 1-\pi_n(x))$ in the
  integral~\eqref{eq:Bayes-opt-measure} to define the measure. %
  In particular, we can construct a \emph{variance}-based measure
  $\Hn^v = \int_\X\pi_n(x)(1-\pi_n(x))\, \dx$, and an
  \emph{entropy}-based one
  $\Hn^e = -\int_\X\pi_n(x)\log_2(\pi_n(x))\, \dx -
  \int_\X(1-\pi_n(x))\log_2(1-\pi_n(x))\, \dx$.  %
  We focus in the following on the misclassification-based QSI-SUR
  strategy~\eqref{eq:QSI-SUR-criterion}. %
  A comparative benchmark provided as Supplementary Material shows
  that the other variants yield almost identical results on the four
  examples of \autoref{sec:numerical}.
\end{remark}

\begin{remark}
  It is instructive to compare the different misclassification-based
  and variance-based criteria to those proposed by
  \cite{bect:2012:sur_failureproba}, and to notice the formal
  resemblance, if replacing~$\pi_n$ by~$p_n$ and~$\Gamma(\xi)$
  by~$\Lambda(\xi)$.
\end{remark}

\subsection{Approximation of the criterion}
\label{sec:approximation}

It appears from the definition of~$\pi_n(x)$ that the proposed
criterion does not admit an explicit form, and thus must be approximated. %
Indeed, it is based on the conditional distribution of~$\tau(x)$,
which is intractable to the best of our knowledge.

In particular, two major issues arise in the numerical evaluation of
$J_n(\Xc,\,\Sc)$ at a given point $(\Xc, \Sc) \in \X\times\S$---namely,
the evaluation of the integral over~$\X$ and the lack of closed-form
formula for the integrand
\begin{equation}
  \label{eq:integrandSUR}
  \En\left[
    \min(\pi_{n+1}(x),1-\pi_{n+1}(x))
    \mid
    (X_{n+1}, S_{n+1}) = (\Xc, \Sc)
  \right].
\end{equation}

To tackle these issues, we propose an approximation of the criterion
based on two ingredients. %
First, using a suitable auxiliary sampling density, the integral
over~$\X$ is estimated using importance sampling. %
Second, at a given point~$x$, the integrand~\eqref{eq:integrandSUR} is
approached using Monte Carlo simulations of conditional sample paths
of the process~$\xi$.

The interested reader can refer to \autoref{SM:sec:criterion-approx} 
for details on this approximation scheme.

\section{Numerical experiments}
\label{sec:numerical}

\subsection{Implementation of the QSI-SUR strategy}
\label{sec:implementation_QSI}

\emph{Bayesian model:} %
The underlying function is modeled using a GP~prior with a constant
mean function and an anisotropic Matérn covariance function. %
The parameters of the GP~prior are estimated, at each step, using the
restricted maximum likelihood (ReML) method \citep[see,
e.g.,][]{stein1999interpolation}, with the constraint that the
regularity parameter $\upsilon$ of the kernel should belong
to~$\left\{\frac{1}{2}, \frac{3}{2}, \frac{5}{2}, +\infty \right\}$. %
Note that the limit case $\upsilon \rightarrow +\infty$ corresponds to
the Gaussian kernel. %
For numerical purposes, in order to limit the occurrence of
ill-conditioned covariance matrices, a nugget of value~$10^{-6}$ is
added. %
Regarding the initial training points, we use a pseudo\footnote{We
  call "pseudo-maximin" the best LHS, in the sense of the minimal
  distance between points, in a collection of 1000 independent
  LHSs.}-maximin LHS, following the rule of thumb which consists in
taking an initial design of size $n_0 = 10d$ \citep[see,
e.g.,][]{loeppky2009choosing}, where $d = d_\X+d_\S$ is the dimension
of the input space.

\emph{Construction of approximation grid:} %
At each step, we first sample $100$ points according to $\Ps$ as a
discretization grid $\tS_n$ of $\S$. Then, a set of
$n_\X = 500\, d_\X$ points is uniformly sampled in $\X$. %
For each of these points, we approximate the misclassification
probability $\min(\pi_n(x), 1-\pi_n(x))$ using Monte Carlo simulations
of $\xi(x,\cdot)$ on $\tS$ (with respect to $\Pn$). %
Finally, a subset $\tX_n$ of $n_{\tX} = 40$~points is constructed,
composed of the point with the highest misclassification probability,
and $n_{\tX} - 1 = 39$~points drawn (without replacement) according to
the discrete probability distribution
$p_\X(x) \propto \min(\pi_n(x), 1-\pi_n(x))$. %
Note that this distribution is chosen to ensure that the elements of
$\tX_n$ are concentrated in the areas of~$\X$ where the probability of
misclassification is high. %
This procedure gives us a product set $\tX_n\times\tS_n$, used to
approximate the integrals involved in the QSI-SUR sampling
criterion. %
The integrand~\eqref{eq:integrandSUR} arising in the criterion is
approximated using a Gauss-Hermite quadrature of $10$~points coupled
with $100$~conditioned sample paths. (See \autoref{SM:sec:criterion-approx} 
for more details on the procedure used to approximate the criterion).

\emph{Optimization of the criterion:} %
The approximated sampling criterion is then optimized using an
exhaustive search over a subset~$\tD_n \subset \tX_n \times \tS_n$ of $250$
candidate points. %
To construct this subset of candidate points, we follow the same idea
that for the construction of $\tX$, using this time sampling probabilities
proportional to the probability of misclassification
$\min(p_n(x,s), 1 - p_n(x,s))$ of $\xi(x,s) \in C$.%

\begin{figure}[tbp]
  \psfrag{0}     {\tiny $0$}
  \psfrag{0.005} {\tiny $0.005$}
  \psfrag{0.01}  {\tiny $0.01$}
  \psfrag{0.015} {\tiny $0.015$}
  \psfrag{0.02}  {\tiny $0.02$}

  \psfrag{2}{\tiny $2$}
  \psfrag{4}{\tiny $4$}
  \psfrag{6}{\tiny $6$}
  \psfrag{8}{\tiny $8$}
  \psfrag{5}{\tiny $5$}
  \psfrag{10}{\tiny $10$}
  \psfrag{15}{\tiny $15$}

  \includegraphics[width=65mm,height=60mm]{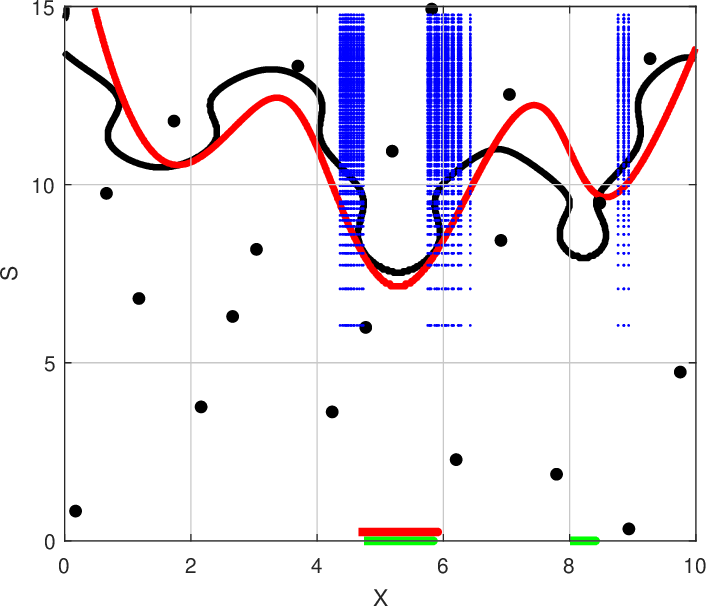} \hspace*{2mm}
  \includegraphics[width=72mm,height=60mm]{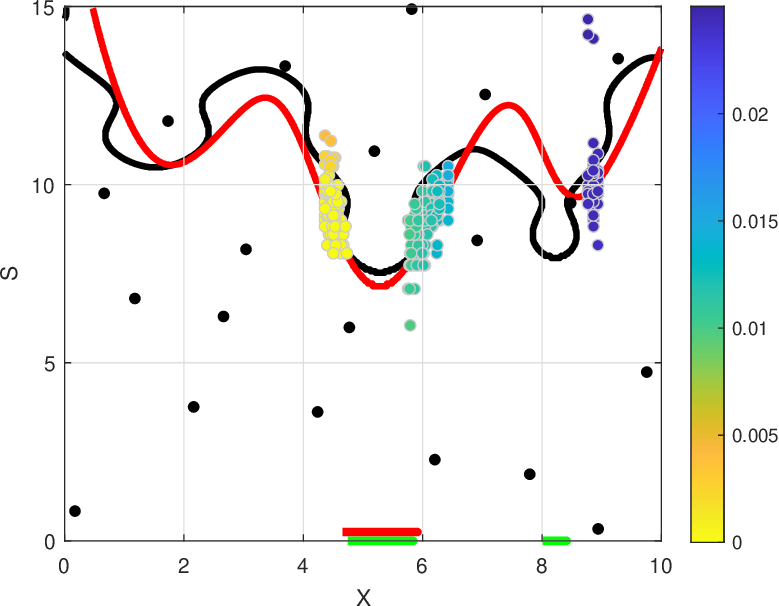}

  \caption{%
    Examples of approximation grid $\tX_n\times\tS_n$ (left) and
    QSI-SUR criterion values on $\tD_n \subset \tX_n\times\tS_n$
    (right), for the acquisition of the first evaluation point
    on~$f_1$. %
    Black dots represent the initial design, the red and black curves
    respectively the estimated and true boundary of $\Lambda(f)$. %
    On the $x$-axis, the segments represent the currently estimated
    quantile set (red), and the true $\Gf$ (green).%
  }%

  \label{fig:optim_illustrated}
  \end{figure}

The main ingredients of this implementation are illustrated in
\autoref{fig:optim_illustrated}.%

\begin{remark}
  \label{rmk:expensive}
  Due to the expensive nature of the simulation of conditional~GPs,
  the simulation parameters described in this section must be
  calibrated according to the computational budget at hand. %
  In particular $\tX_n$ and $\tS_n$ must be constructed such that
  $\card(\tX_n \times \tS_n)$ is sufficiently low to accommodate a given
  computation time.
\end{remark}

\subsection{Comparative methods and performance metric}
\label{sec:comp_meth}

Due to the absence in the literature of strategies tailored specifically to the
QSI problem, we propose to compare the performance of the
QSI-SUR criterion against methods that aim at approximating the
set~$\Lambda(f)$ in the joint space~$\X \times \S$.

Using the same Bayesian modeling as for the QSI-SUR criterion, the
results are first compared to the SUR criterion of
\cite{bect:2012:sur_failureproba}, hereafter denoted as ``Joint-SUR'',
using the closed-form expression of
\cite{chevalier:2014:fast_parallel_kriging}. %
We draw $5d_\X\times10^4$ points according to the uniform distribution
on $\X\times\S$. %
The criterion is approximated using a subset of $4000$ points
constructed from the points with the highest misclassification
probability $\min(p_n(x,s), 1-p_n(x,s))$ and a sample (without
replacement) according to this probability. %
It is then optimized using an exhaustive search on $250$ of those
points.

We also compare against the criterion of \citet{ranjan:2008:contour}
with $\kappa = 1.96$ and the misclassification probability criterion of
\cite{bryan:2005:active}, both evaluated on $5d_\X\times10^4$ points
sampled uniformly on $\X\times\S$. %

The Entropy Contour Locator (ECL) criterion of
\cite{cole:2023:entropy} is also considered, with $4000$ candidate
points, using the Python
implementation\footnote{\url{https://bitbucket.org/gramacylab/nasa/src/master/}}
provided by the authors, with minor modifications to fit our Bayesian
modeling. %
As explained in \autoref{sec:pointwise}, ECL and the misclassification
probability criterion are equivalent in principle: %
the main difference lies in the use, in ECL, of a local optimizer from
the best candidate point %
(hence the use of a smaller number of candidate points).

Finally, as a baseline, we include the results obtained by
uniform random sampling on the space $\X\times\S$.

Except for the ECL criterion, all the experiments are carried out
using Matlab R2022a and the STK toolbox v2.8.1 \citep{STK}. %
The implementation of the
methods\footnote{\url{https://github.com/stk-kriging/contrib-qsi}}
(except ECL) and the scripts used to run the numerical
experiments\footnote{\url{https://github.com/stk-kriging/qsi-paper-experiments}}
are available online. %

To assess the performances of the methods, we compare at each step the
proportions of misclassified points obtained on a prediction grid
composed of the product of a Sobol' sequence of $2^{12}$ points
in~$\X$ and the inversion (with respect to the cumulative distribution
function of $\Ps$) of a Sobol sequence of $2^{10}$ points in
$[0,1]^{d_\S}$. %
Considering that the Bayes-optimal
estimator~\eqref{eq:Bayes-opt-estim} is expensive to approximate on
such a large grid, we use instead the estimator
$\hGn = \{x \in \X \, : \, \En(\tilde\tau(x)) \le \alpha\}$, %
where $\tilde\tau(x)$ is the approximation of $\tau(x)$ defined by
averaging over the $2^{10}$ selected points of~$\S$.

Each method is run $100$ times on each test case, using different
initial designs, to study performances variability.

\subsection{Synthetic examples}
\label{sec:presentation}

We propose first three synthetic examples, with scalar output values
($q = 1$) and noise-free observations.

The first test function (see \autoref{fig:f1}), defined
on~$\X = [0; 10]$ and~$\S = [0; 15]$, is a modified Branin-Hoo
function
$f_1(x,s) = \frac{1}{12}\, b(x,s) + 3\sin\left(x^{\frac{5}{4}}\right)
+ \sin\left(s^{\frac{5}{4}}\right)$, where $b$ is the Branin-Hoo
function
$b(x,s) = \left(s - \frac{5.1x^2}{4\pi^2} + \frac{5x}{\pi} - 6
\right)^2 + 10 \left(1-\frac{1}{8\pi}\right)\cos(x) + 10$
\citep{branin:1972:branin-hoo}. %
We take $C = (-\infty; T]$ with $T = 7.5$, $\alpha = 0.05$ and
$\Ps$~the Beta distribution with parameters $(7.5, 1.9)$, rescaled
from~$\left[ 0, 1 \right]$ to~$\S$. %
The associated set~$\Gamma(f_1)$ is represented in~\autoref{fig:f1}.

The test function $f_2$ is defined by
$f_2\left( x,s \right) = \frac{1}{2}\, c(x_1,s_1) + \frac{1}{2}\,
c(x_2,s_2)$, on~$\X = [-2;2]^2$ and~$S = [-1;1]^2$, with $c$ the
"six-hump camel" function
$c(x,s) = \left(4-2.1x^2+\frac{x^4}{3}\right)x^2 + xs + (4s^2-4)s^2$
\citep{dixon:1978:intro}. %
We take $C = (-\infty; T]$ with $T = 1.2$, $\alpha = 0.15$ and
$\Ps$~the uniform distribution on~$\S$.

As a third test function $f_3$, we consider the Hartman4 function
\citep{picheny:2013:benchmark}, defined on~$\X = [0;1]^2$
and~$\S = [0;1]^2$ equipped with the uniform distribution. %
For this example we set $\alpha = 0.6$ and $C = [T, +\infty)$, with
$T = -1.1$.

\begin{figure}
  \centering
  \includegraphics[width=\toto]{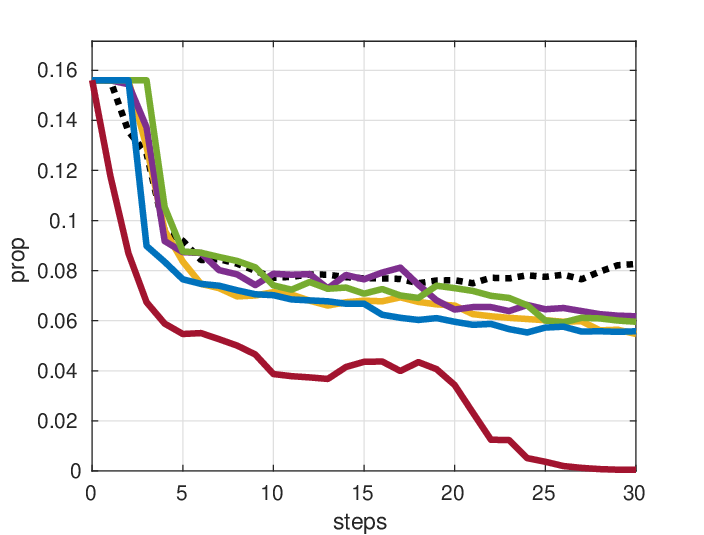}
  \hspace*{5mm}
  \includegraphics[width=\toto]{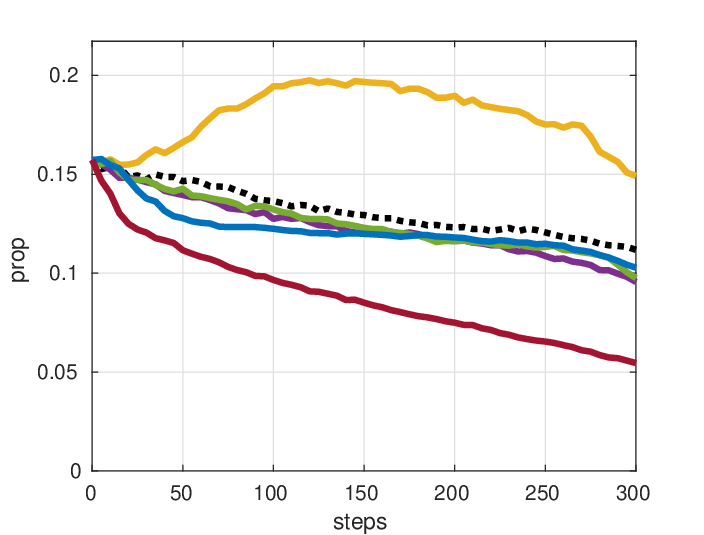}\\[5mm]
  \includegraphics[width=\toto]{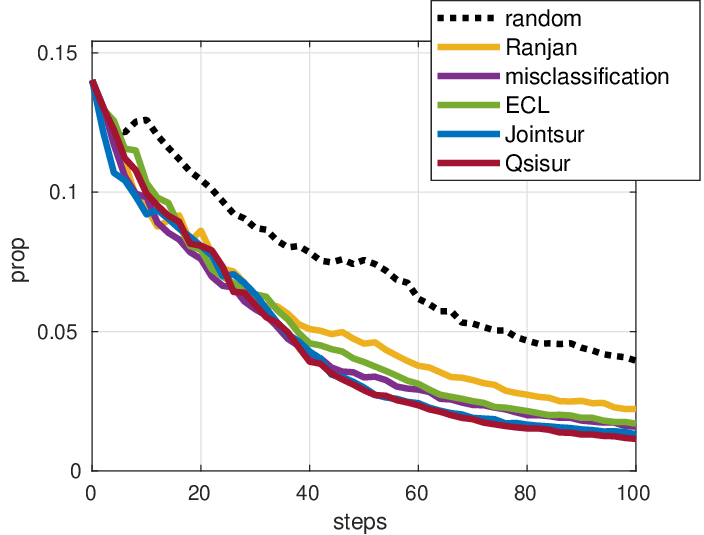}
  \caption{%
    Median of the proportion of misclassified points vs.\;number of
    steps, for 100~repetitions of the algorithms on the test
    functions~$f_1$ (top left), $f_2$ (top right) and~$f_3$
    (bottom). }
  \label{fig:results_f1_f2}
\end{figure}

From the median performance results in \autoref{fig:results_f1_f2}, it
can be observed that the new sampling criteria tends to perform better
on the QSI problem than the state-of-the-art methods focusing on the
set~$\Lambda(f)$ in the joint space $\X \times \S$. %
More specifically, it performs much better on the first two cases, and
has similar performance on the third one. %
This remains true when looking at the 75th and 95th percentiles, as
illustrated in \autoref{fig:results_f1_quant} for~$f_1$ (and in the
Supplementary Material for the other cases).

\begin{figure}
  \centering
  \includegraphics[width=\toto]{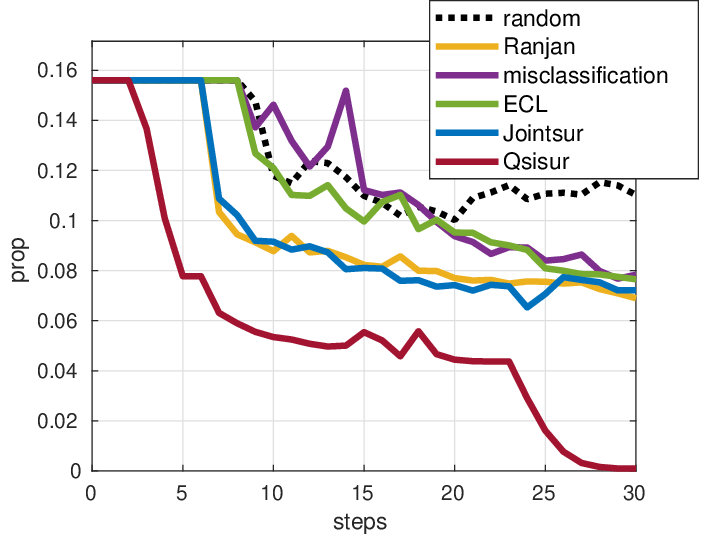}
  \hspace*{5mm}
  \includegraphics[width=\toto]{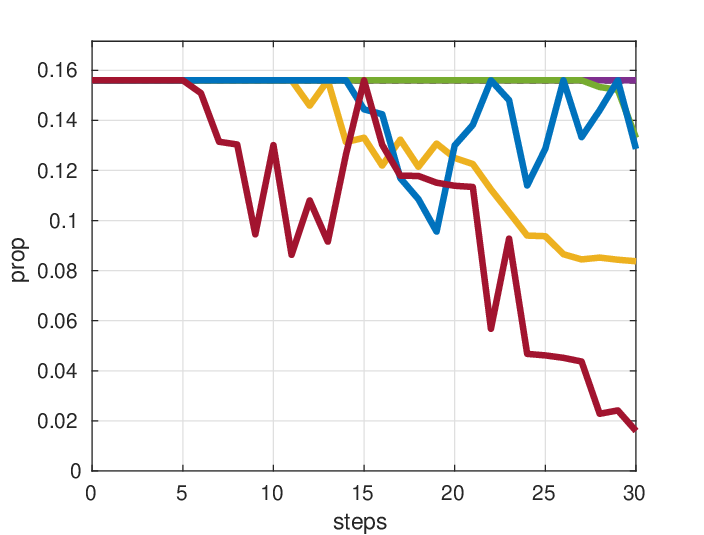}
  \caption{Quantiles of order $0.75$ (left) and $0.95$ (right) of the
    proportion of misclassified points vs.\;number of steps, for
    100~repetitions of the algorithms on the test function~$f_1$. }
  \label{fig:results_f1_quant}
\end{figure}

\begin{figure}
  \centering
  \psfrag{0}{\tiny $0$}
  \psfrag{0.1}{\tiny $0.1$}
  \psfrag{0.2}{\tiny $0.2$}
  \psfrag{0.4}{\tiny $0.4$}
  \psfrag{0.6}{\tiny $0.6$}
  \psfrag{0.8}{\tiny $0.8$}
  \psfrag{1}{\tiny $1$}
  \psfrag{2}{\tiny $2$}
  \psfrag{-2}{\tiny $-2$}
  \psfrag{4}{\tiny $4$}
  \psfrag{6}{\tiny $6$}
  \psfrag{8}{\tiny $8$}
  \psfrag{5}{\tiny $5$}
  \psfrag{10}{\tiny $10$}
  \psfrag{12}{\tiny $12$}
  \psfrag{14}{\tiny $14$}
  \psfrag{15}{\tiny $15$}
  \psfrag{16}{\tiny $16$}
  \psfrag{18}{\tiny $18$}
  \psfrag{Ranjan}[c][c]{\small \raisebox{3pt}{Ranjan}}
  \psfrag{misclassification}[c][c]{\small \raisebox{3pt}{misclassification}}
  \psfrag{QSI}[c][c]{\small \raisebox{3pt}{QSI-SUR}}
  \psfrag{joint}[c][c]{\small \raisebox{3pt}{Joint-SUR}}
  \includegraphics[width=\toto]{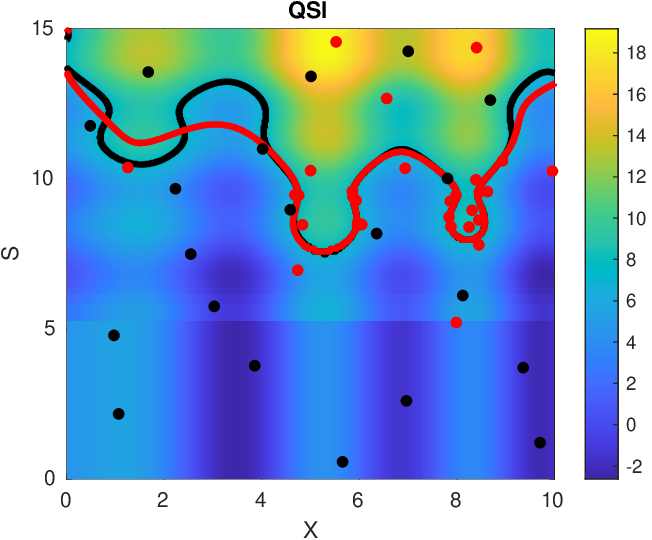} \hspace*{5mm}
  \includegraphics[width=\toto]{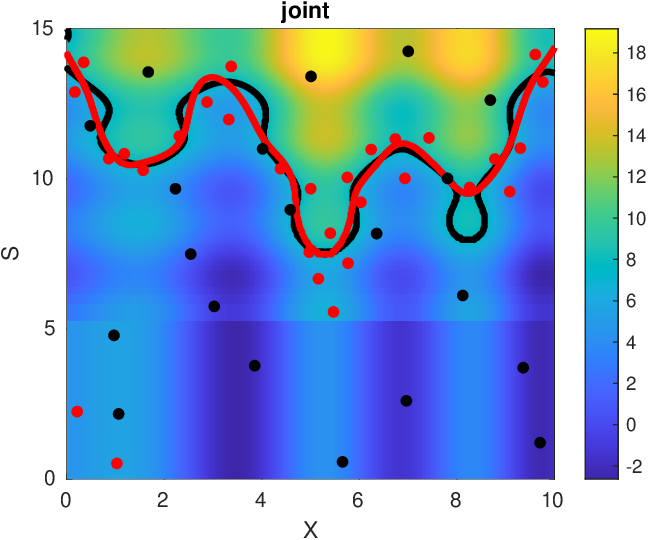}\\[5mm]
  \includegraphics[width=\toto]{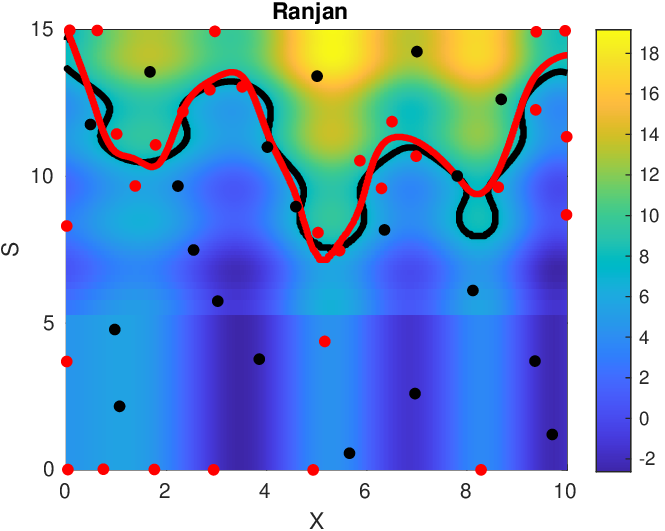} \hspace*{5mm}
  \includegraphics[width=\toto]{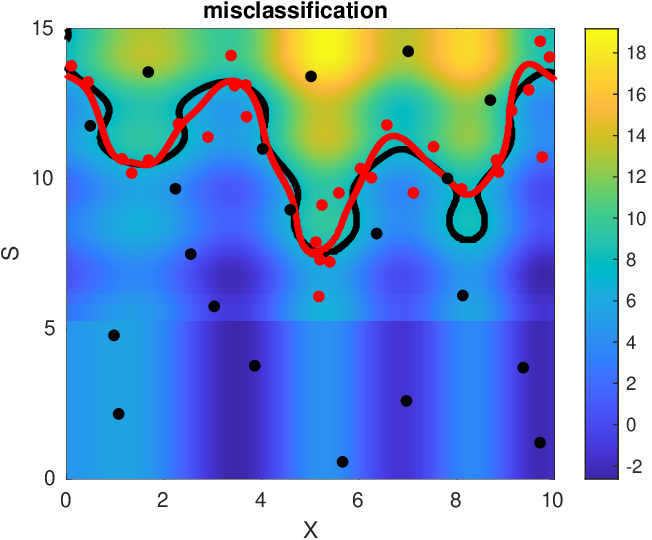}
  \caption{Examples of sequential designs (red dots) obtained after
    $30$~steps on the function~$f_1$, with the QSI-SUR and three others
    sampling criteria. Black dots represent the initial design points,
    black curve the boundary of $\Lambda(f)$, and red curve the
    boundary of $\Lambda(\mn)$.}
  \label{fig:design_f1}
\end{figure}

\begin{figure}

  \psfrag{Ranjan}[c][c]{\small \raisebox{3pt}{Ranjan}}
  \psfrag{misclass}[c][c]{\small \raisebox{3pt}{misclassification}}
  \psfrag{Qsi}[c][c]{\small \raisebox{3pt}{QSI-SUR}}
  \psfrag{joint}[c][c]{\small \raisebox{3pt}{Joint-SUR}}

  \psfrag{0}[c][c]{\tiny $0$}
  \psfrag{0.005} {\tiny $0.005$}
  \psfrag{0.01}  {\tiny $0.01$}
  \psfrag{0.015} {\tiny $0.015$}
  \psfrag{0.02}  {\tiny $0.02$}
  \psfrag{0.025} {\tiny $0.025$}
  \psfrag{0.05}  {\tiny $0.05$}
  \psfrag{0.1}   {\tiny $0.1$}
  \psfrag{0.15}  {\tiny $0.15$}
  \psfrag{0.188} {\tiny $0.188$}
  \psfrag{0.19}  {\tiny $0.19$}
  \psfrag{0.192} {\tiny $0.192$}
  \psfrag{0.194} {\tiny $0.194$}
  \psfrag{0.196} {\tiny $0.196$}
  \psfrag{0.198} {\tiny $0.198$}
  \psfrag{0.2}   {\tiny $0.2$}
  \psfrag{0.202} {\tiny $0.202$}
  \psfrag{0.25}  {\tiny $0.25$}
  \psfrag{0.3}   {\tiny $0.3$}
  \psfrag{0.35}  {\tiny $0.35$}
  \psfrag{0.4}   {\tiny $0.4$}
  \psfrag{0.45}  {\tiny $0.45$}

  \psfrag{1}{\tiny $1$}
  \psfrag{2}{\tiny $2$}
  \psfrag{3}{\tiny $3$}
  \psfrag{4}{\tiny $4$}
  \psfrag{5}[c][c]{\tiny $5$}
  \psfrag{6}{\tiny $6$}
  \psfrag{7}{\tiny $7$}
  \psfrag{8}{\tiny $8$}
  \psfrag{9}{\tiny $9$}
  \psfrag{10}[c][c]{\tiny $10$}
  \psfrag{12}{\tiny $12$}
  \psfrag{15}[c][c]{\tiny $15$}
  \psfrag{20}{\tiny $20$}
  \psfrag{25}{\tiny $25$}
  \psfrag{30}{\tiny $30$}
  \psfrag{35}{\tiny $35$}
  \psfrag{40}{\tiny $40$}
  \psfrag{45}{\tiny $45$}

  \includegraphics[width=45mm]{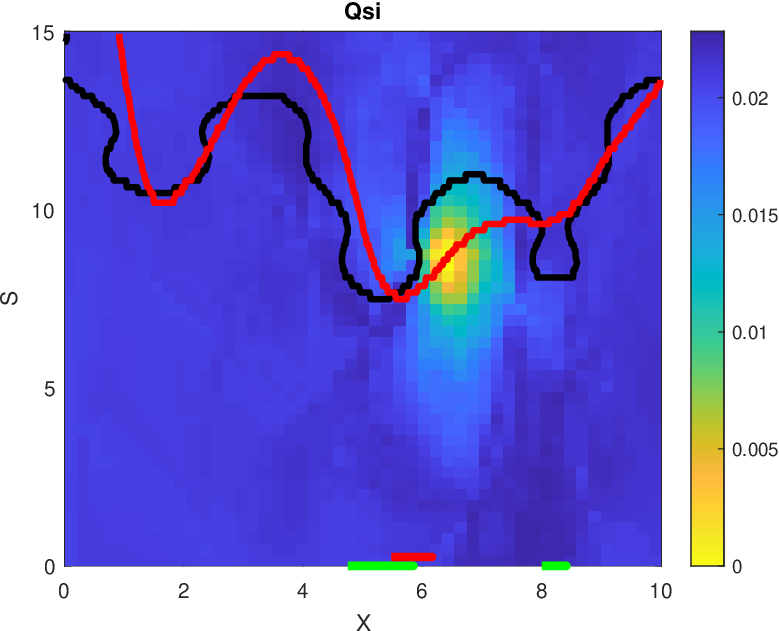}\hspace{2mm}
  \includegraphics[width=45mm]{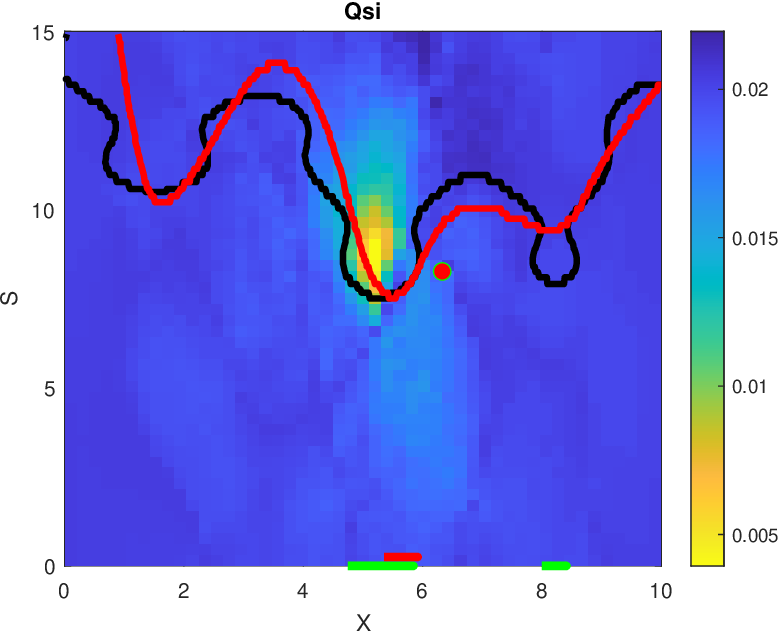}\hspace{2mm}
  \includegraphics[width=45mm]{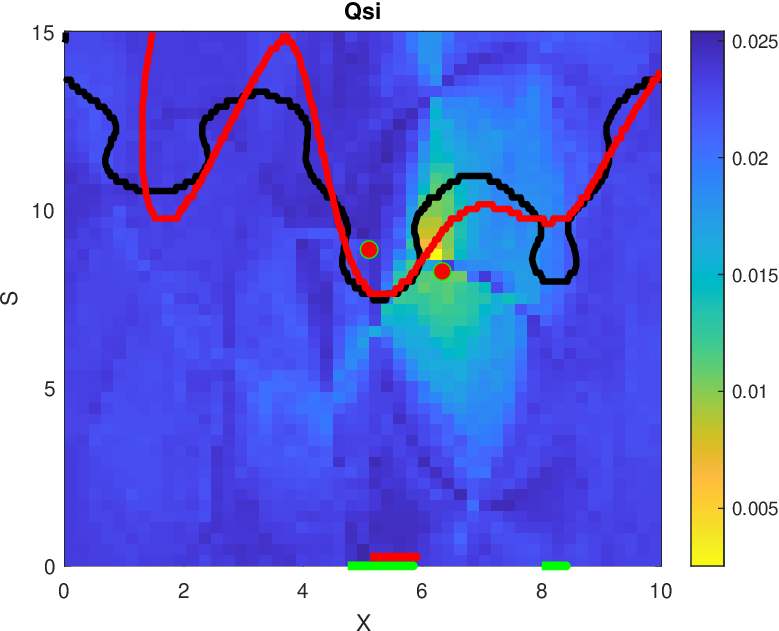} \\[5mm]
  \includegraphics[width=45mm]{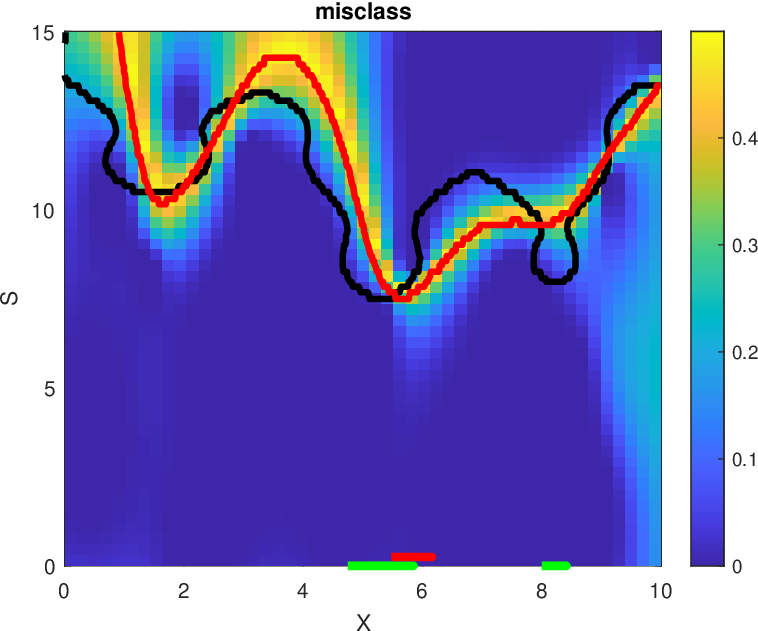}\hspace{2mm}
  \includegraphics[width=45mm]{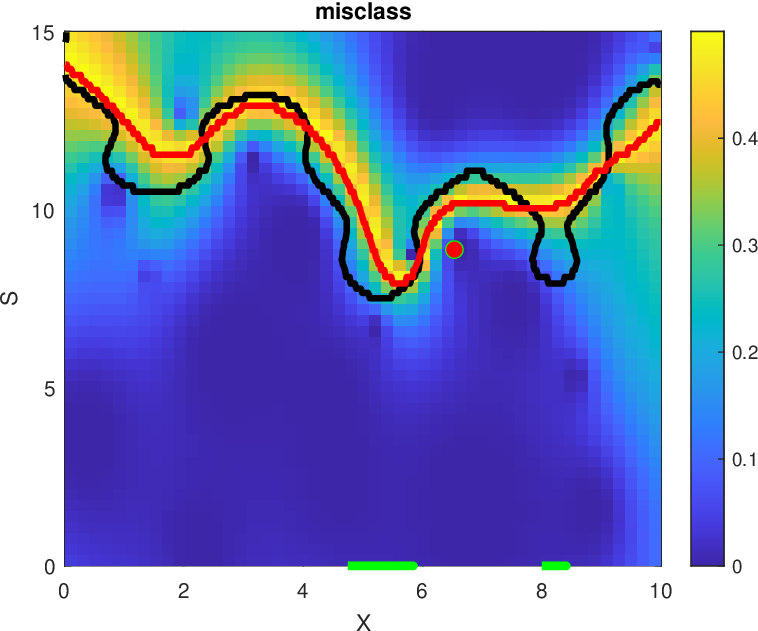}\hspace{2mm}
  \includegraphics[width=45mm]{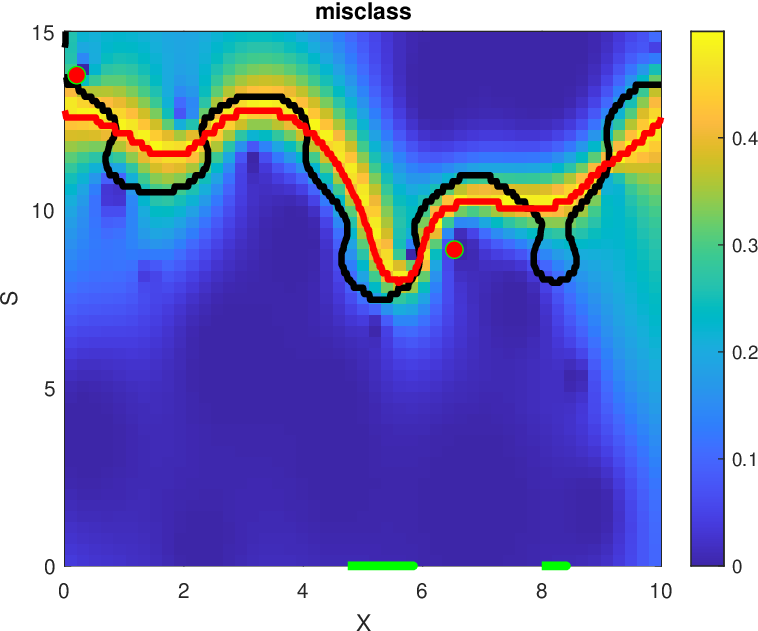}
  \caption{%
    Examples of values of the QSI-SUR (top) and maximum
    misclassification probability (bottom) criteria, for step~$1$
    to~$3$ (left to right), for the function~$f_1$. %
    The black curve represents the boundary of $\Lambda(f)$, and the
    red curve its estimation. Red dots are the points evaluated at the
    previous steps. %
    The segments on the $x$-axis represent $\Gf$ (green) and its
    estimation (red). %
    Evaluated on a grid of $50\times50$ points.%
  }%

  \label{fig:surface_criterion}
  \end{figure}

These differences in performance can be explained, from a heuristic
viewpoint, by the fact that, in the joint space $\X\times\S$, %
the new criterion tend to concentrate the evaluations around specific
zones of~$\Lambda(f)$ which are particularly relevant for the
approximation of~$\Gamma(f)$. %
This phenomenon, related to both the geometry of~$\Lambda(f)$ and the
distribution $\Ps$, can be visualized in \autoref{fig:design_f1}
for~$f_1$. %
In relation to the sequential designs displayed in
\autoref{fig:design_f1}, it can be observed in
\autoref{fig:surface_criterion} that the competitor method (here,
maximum misclassification probability criterion) tends to select
points that are close to the estimated boundary of $\Lambda(f)$,
or/and have high variance. %
In comparison, the QSI-SUR strategy focuses mainly on areas of the
joint space $\X\times\S$ that are susceptible---given the current
data---to provide information about $\Gamma(f)$. %
It is important to notice, however, that in some cases (illustrated
here by the function $f_3$), the performances of the two kinds of
methods are similar.

This gain in performance has to be put in perspective with the higher
computation time of the proposed method, %
which is, depending on the test case, between 5 and 25 times slower
than the simplest methods (e.g., the method of
\cite{ranjan:2008:contour}, see \autoref{SM:sec:details-cost}). %
It is only slightly more expensive, however, than the ``Joint-SUR''
method (approximately twice slower in our benchmark). %
This higher computation time remains quite acceptable for expensive
computer models, but simpler methods remain relevant for moderately
expensive ones (taking, e.g., a few minutes per run).

\subsection{Application to history matching}

The objective of this application is to retrieve the set of plausible
deterministic input variables of a numerical simulator, given
real-life measurements. %
Such a problem can be seen as a particular case of a "history matching"
problem \citep[see, e.g.,][]{williamson:2013:history_matching}. %
More precisely, we consider an uncertain Mogi model
\citep{mogi:1958:volcano}, which simulates the displacement
at the surface of a volcano caused by an underground magma reservoir,
while taking into account the mechanical property of the soil
\citep{durrande:2017:lecture_gp}.

Formally, the model can be seen as a function
$v : \X\times\S \to \Rset^{220}$, with inputs
$(x_1, \ldots, x_5) \in \X = [0,1]^5$ representing the
normalized latitude, longitude, elevation, radius and overpressure of
the magma source, and $(s_1, s_2) \in \S = [0,1]^2$ representing
uncertain perturbations of the shear modulus $G$ and Poisson ratio
$\nu$ of the material, which are written as %
\begin{align*}
  G(s_1) = 2000+100(2s_1-1) \quad \text{and} \quad \nu(s_2) = 0.25+0.3(2s_2-1).
\end{align*}  %
The uncertain variables are assumed independent and identically
distributed, following a Beta distribution with parameters~$(2,2)$.

\begin{figure}
  \centering
  \psfrag{latitude}[c][c]{\raisebox{-10pt}{latitude}}
  \psfrag{longitude}[c][c]{\raisebox{10pt}{longitude}}
  \includegraphics[width=10cm]{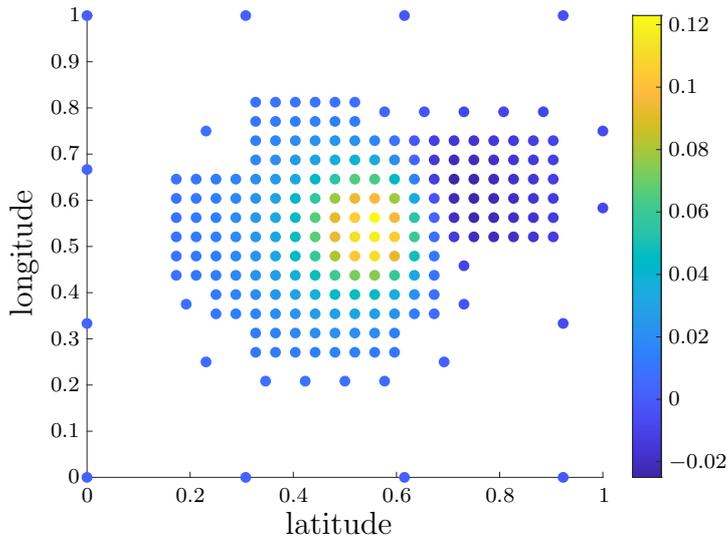}
  \caption{%
    True measures of the displacement at the volcano surface with
    respect to the normalized latitude and longitude.}
  \label{fig:true_measures_volcano}
\end{figure}

Given the real measurements $(y_i)_{i \in \{1,\,...\,,220\}}$ of the
displacement at the surface of the volcano (illustrated
in~\autoref{fig:true_measures_volcano}) and considering the mean
absolute error of the simulated displacement against the real measures
$V(x,s) = \frac{1}{220}\sum_{i=1}^{220}|v_i(x,s)-y_i|$, our objective
is to retrieve the set of plausible parameters for the Mogi model.
More specifically, a vector of parameters
$x \in \X$ is considered to be "plausible" if it yields an error~$V(x, S)$
strictly less than~$0.015$ with a probability larger than~$10\%$. %
To exhibit the direct link between this history matching problem and
the QSI framework, notice that it can be equivalently reformulated as
the problem of estimating the set $\Gamma(V)$, with critical region
$C = [0.015, +\infty)$ and $\alpha = 90\%$.

\begin{figure}[h]
  \centering
  \includegraphics[width=\toto]{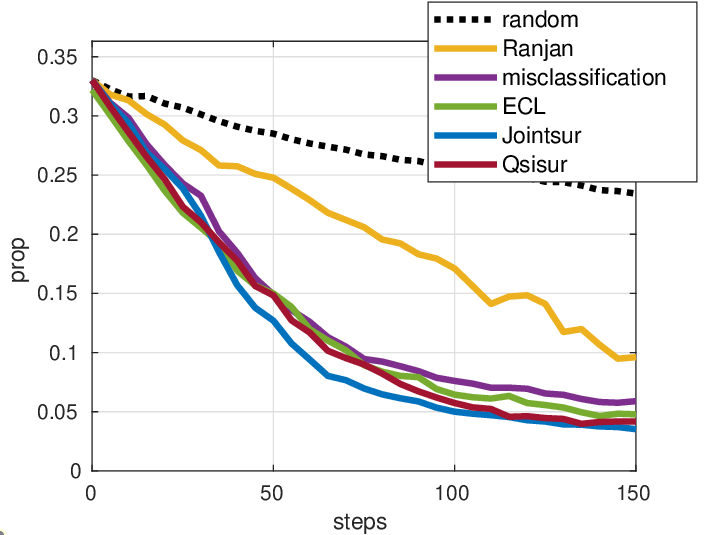}
  \hspace*{5mm}
  \includegraphics[width=\toto]{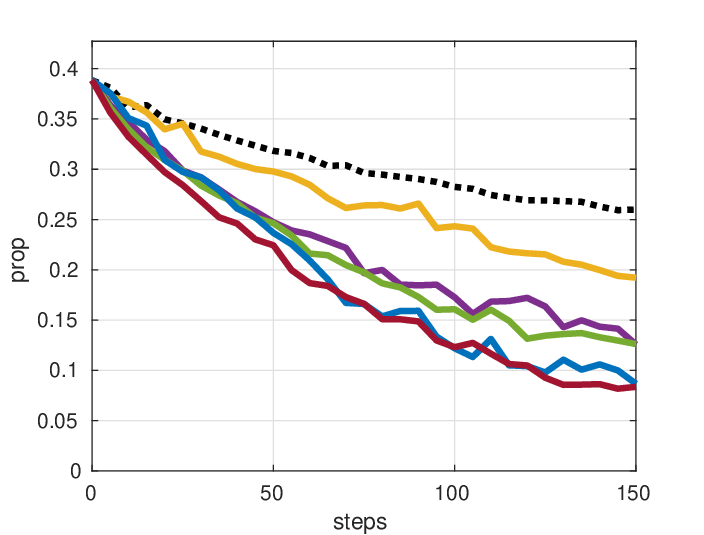}
  \caption{%
    Median (left) and quantile of order $0.95$ (right) of the
    proportion of misclassified points vs.\;number of steps, for
    100~repetitions of the algorithms on "volcano" test case.}
  \label{fig:results_volcano}
\end{figure}

We observe in \autoref{fig:results_volcano} that the median proportion
of misclassified points decreases similarly to the top competitors
(namely, ECL, probability of misclassification and "Joint-SUR"), with
approximately equal median performances after $150$~steps. %
However, our strategy proposes the best "worst case" results, as
indicated by the quantile of order $0.95$ of the proportion of
misclassified points.

\section{Conclusion}
\label{sec:conclusion}

This article presents a SUR strategy for a particular set inversion
problem, in a framework where a function admits deterministic and
uncertain input variables, that we called Quantile Set Inversion
(QSI). %
The practical interest of the proposed method is illustrated on
several problems, on which methods that do not take advantage of the
specificity of the QSI problem tend to be outperformed. %
However, this gain in performance comes at the cost of a high
numerical complexity, in relation to the heavy use of conditioned
Gaussian trajectory simulations. %
Future work will concentrate on reducing this numerical complexity
and making the method applicable to harder
test problems, notably in the case of high-dimensional inputs and
small quantile sets $\Gf$. %
In an other direction, the proposed method could benefit from some adaptations
to make it more applicable to real-life problem---in particular, its adaptation
to batch design in order to tackle cases
where several instances of the simulator can be run in parallel. %

\paragraph{Acknowledgments.} %
The authors are grateful to Rodolphe Le Riche and Valérie Cayol
for sharing their R implementation of the
Mogi model used in \autoref{sec:numerical}.

\bibliographystyle{chicago}
\bibliography{qsi-refs}

\cleardoublepage\appendix
\newcounter{main_figure}
\setcounter{main_figure}{\value{figure}}
\renewcommand\thefigure   {SM\fpeval{\value{figure}-\value{main_figure}}}

\newcounter{main_equation}
\setcounter{main_equation}{\value{equation}}
\renewcommand\theequation   {SM\fpeval{\value{equation}-\value{main_equation}}}

\newcounter{main_table}
\setcounter{main_table}{\value{table}}
\renewcommand\thetable   {SM\fpeval{\value{table}-\value{main_table}}}

\newcounter{main_remark}
\setcounter{main_remark}{\value{remark}}
\renewcommand\theremark   {SM\fpeval{\value{remark}-\value{main_remark}}}

\setcounter{section}{0}

\noindent\rule{\textwidth}{4pt}
\section*{\centering SUPPLEMENTARY MATERIAL}
\addtocontents{toc}{\protect\vspace{20pt}}
\addcontentsline{toc}{section}{SUPPLEMENTARY MATERIAL}
\vspace{5pt}
\rule{\textwidth}{4pt}
\vspace{5mm}

\psfrag{bigbigbigquan}{\small $5\%$ - $95\%$}
\psfrag{bigbigbigquan2}{\small $25\%$ - $75\%$}
\psfrag{medianmedian}{\small median}

\section{\texorpdfstring{Proof of the expression of $\Hn$}{Proof of the expression of Hn}}
\label{SM:sec:proof-Hn}

Let us remark that
\begin{equation*}
  \Gxi\Delta\hGn = \left\{x \in \Gxi \, : \, x \notin \hGn\right\}\bigcup\left\{x \in \hGn \, : \, x \notin \Gxi\right\}.
\end{equation*}
As a consequence, by defining the classifier
$c_n(x) = \mathds{1}_{\hGn}(x)$ and by Fubini's theorem:
\begin{align*}
  \En\left[\lambda\left(\Gxi\Delta\hGn\right)\right]
  & = \int_\X\En\left[\mathds{1}_{\Gxi\Delta\hGn}(x)\right]\, \dx\\
  & = \int_\X \En\left[\mathds{1}_{\{c_n = 0\}}(x)\mathds{1}_\Gxi(x)\right]\\
  & \qquad\qquad + \int_\X \En\left[\mathds{1}_{\{c_n = 1\}}(x)\left(1-\mathds{1}_\Gxi(x)\right)\right]\, \dx\\
  & = \int_\X\mathds{1}_{\{c_n = 0\}}(x)\, \pi_n(x)\, \dx\\
  & \qquad\qquad + \int_\X\mathds{1}_{\{c_n = 1\}}(x)\left(1-\pi_n(x)\right)\, \dx.
\end{align*}
It suffices to observe that, if
$\hGn = \left\{x\in\X \, : \, \pi_n(x)>\frac{1}{2}\right\}$, then
$c_n(x) = \mathds{1}_{\left\{\pi_n(x) > \frac{1}{2}\right\}}(x)$, and
for all $x \in \X$:
\begin{align*}
  \mathds{1}_{\{c_n = 0\}}(x)\pi_n(x) + \mathds{1}_{\{c_n = 1\}}(x) & \left(1-\pi_n(x)\right)\\
  & = \mathds{1}_{\left\{\pi_n \le \frac{1}{2} \right\}}(x)\pi_n(x) + \mathds{1}_{\left\{\pi_n > \frac{1}{2} \right\}}(x)\left(1-\pi_n(x)\right)\\
  & = \min\left(\pi_n(x), 1-\pi_n(x)\right)
\end{align*}
to obtain the simplified expression of $\Hn$.

\section{Approximation of the criterion}
\label{SM:sec:criterion-approx}

We give here the details about the approximation of the criterion
$J_n(\Xc,\,\Sc)$. %
The same procedure can be adapted for the variations of the criterion
based on the variance or the entropy.

The integral on $\X$ (with respect to the uniform distribution) is
estimated using an importance sampling scheme. %
This allows to non-uniformly sample an approximation grid~$\tX$ for
our integral in order, for instance, to concentrate the sampled points
in uncertain areas of~$\Gxi$. %
Given a random finite collection $\tX$ of elements of $\X$ sampled
from a density $p_\X$, we use the following importance sampling
approximation:
\begin{equation}
  \label{eq:IS_approx}
  J_n(\Xc,\,\Sc) \;\approx\; \sum_{x \in \tX}\frac{1}{p_\X(x)}\En\left[
    \min(\pi_{n+1}(x),1-\pi_{n+1}(x))
    \mid
    (X_{n+1}, S_{n+1}) = (\Xc, \Sc)
  \right].
\end{equation}

We propose to estimate the integrand~\eqref{eq:integrandSUR} using
quantization of the distribution~$\Ps$ together with Monte Carlo
simulations of the process~$\xi$, %
in the spirit of~\cite{villemonteix:2009:informational}.

Consider a finite subset $\tS$ of~$\S$, and a family
$\left( w_\S(s) \right)_{s \in \tS}$ of positive real numbers such
that $\PtildeS = \sum_{s\in\tS}^{}w_\S(s)\delta_{s}$ is a "good"
approximation of~$\Ps$, where $\delta_{s}$ denotes the Dirac measure
at~$s$. %
This can be achieved, for instance (as done in
\autoref{sec:numerical}), by defining $\tS$ as a collection of $n_\S$
i.i.d. samples from $\Ps$ and fixing $w_\S(s) = \frac{1}{n_\S}$ for
all $s \in \tS$. %
For more information about quantization, the reader can refer
to~\cite{graf:2000:quantization}.

Moreover, let $\{z_1,\dotsc,z_N\}$ and
$(w_\xi(z_i))_{i \in \{1,\,...\,, N\}}$ be such that
$\sum_{i=1,\,...\,, N}w_\xi(z_i)\delta_{z_i}$ is a quantization of the
distribution of $\xi(\Xc,\Sc)$ given $\In$ (for example a
Gauss-Hermite quadrature), and recall that $\tX \subset \X$ is the
finite subset used for the approximation of the integral over~$\X$
arising in $J_n(\Xc,\,\Sc)$. %
$\xi$ being Gaussian, assuming that $\tX\times\tS$ is not too large we
can easily simulate $M$ sample paths
$\{\xi_{i,1}, \ldots, \xi_{i,M}\}$ of $\xi$ over $\tX\times\tS$, under
the distribution $\Pn(\, \cdot \mid \xi(\Xc, \Sc) = z_i)$. %
Given a point $x \in \tX$, set
\begin{align}
  \tilde{\pi}_{n+1}^i(x)
  = \frac{1}{M}\sum_{m=1}^{M}\mathds{1}_{[0,\alpha]}\left(\sum_{s\in\tS}w_\S(s)\, \mathds{1}_C(\xi_{i,m}(x,s)) \right).
  \label{approx-eq}
\end{align}
For a sufficiently large $M$ and a ``good'' quantization~$\PtildeS$, we have
\begin{equation}
  \tilde{\pi}_{n+1}^i(x) \;\approx\;
  \P\left(\tau(x) \le \alpha \mid \In\,, \, \xi(\Xc, \Sc) = z_i\right)\,.
\end{equation}
As a consequence, it is possible to use
\begin{equation}
  \label{equ:approx-integrand}
  j^x_{n}(\Xc, \Sc) = \sum_{i=1}^N w_\xi(z_i)\, \min(\tilde{\pi}_{n+1}^i(x), 1-\tilde{\pi}_{n+1}^i(x))
\end{equation}
as an approximation of~\eqref{eq:integrandSUR}.

Combining~\eqref{eq:IS_approx} and~\eqref{equ:approx-integrand}, the
criterion~$J_n(\Xc, \Sc)$ is then approximated by
\begin{equation}
  \label{eq:approx-SUR}
  \tilde{J}_n(\Xc,\Sc) \;=\; \sum_{x\in\tX} \frac{1}{p_\X(x)}\, j^x_{n}(\Xc, \Sc).
\end{equation}

\begin{remark}
  For a better numerical efficiency, the simulations of the sample
  paths of $\xi$ under $\Pn(\cdot \, | \, \xi(\Xc,\Sc) = Z_i)$ are
  preferably carried out using reconditioning of sample paths. %
  A description of this procedure is given by
  \cite{villemonteix:2009:informational}, Section~5.1.
\end{remark}

\section{Details on computational cost}
\label{SM:sec:details-cost}

Due to the major implementation differences between the Entropy
Contour Locator (ECL) method of \cite{cole:2023:entropy} and the
others competitors, we exclude it of this benchmark. We focus here on
the strategies implemented in Matlab using the STK toolbox v2.8.1
\citep{STK}. These experiments are conducted using Matlab R2022a and
the same parameters as described in \autoref{sec:numerical}, on a
computer equipped with a CPU AMD Ryzen 7 3700x with 32GB of RAM.

\vfill

\begin{table}[ht]
  \centering
\begin{tabular}{l|c | c | c | c|}
  \cline{2-5}
  & Ranjan        & misclass.        & Joint-SUR      & QSI-SUR  \\
  \hline
  \multicolumn{1}{|l|}{$f_1$}    &0.15 & 0.14 & 3.74 &  3.77 \\
  \hline
  \multicolumn{1}{|l|}{$f_2$}    & 0.29 & 0.23 & 7.01  & 5.91   \\
  \hline
  \multicolumn{1}{|l|}{$f_3$}    & 0.30 & 0.23 & 6.57 & 5.35  \\
  \hline
  \multicolumn{1}{|l|}{Volcano}  & 1.24 & 0.76 & 12.77 & 11.30  \\
  \hline
\end{tabular}
\caption{%
  Runtime (in seconds) to complete the first step. %
  Average over $10$~runs.}
\label{table:first_time}
\end{table}

\vfill

\begin{table}[ht]
  \centering
  \begin{tabular}{l|c | c | c | c|}
    \cline{2-5}
    & Ranjan        & misclass.        & Joint-SUR      & QSI-SUR  \\
    \hline
    \multicolumn{1}{|l|}{$f_1$}    &1 & 0.81 & 19.26 &  21.94  \\
    \hline
    \multicolumn{1}{|l|}{$f_2$}    & 1 & 0.84 & 4.96  & 9.45   \\
    \hline
    \multicolumn{1}{|l|}{$f_3$}    & 1 & 0.83 & 10.02 & 11.05  \\
    \hline
    \multicolumn{1}{|l|}{Volcano}  & 1 & 0.53 & 4.26 & 7.19  \\
    \hline
  \end{tabular}
  \caption{%
    Normalize total runtime (in seconds). %
    Average over $10$~runs.}
  \label{table:total_time}
\end{table}

\section{Comparison between variants of the QSI-SUR criterion}
\label{SM:sec:results-variants}

Following~\autoref{rem:other-uncertainty}, we display here a brief comparison
of several variants of the QSI-SUR criterion---namely, the
misclassification probability-based, variance-based, and entropy-based
sampling criteria.

\begin{figure}[ht]

  \psfrag{bigbigmiscbased}{\small misclass.}
  \psfrag{bigbigvarbased}{\small variance}
  \psfrag{bigbigentrbased}{\small entropy}

  \includegraphics[width=\toto]{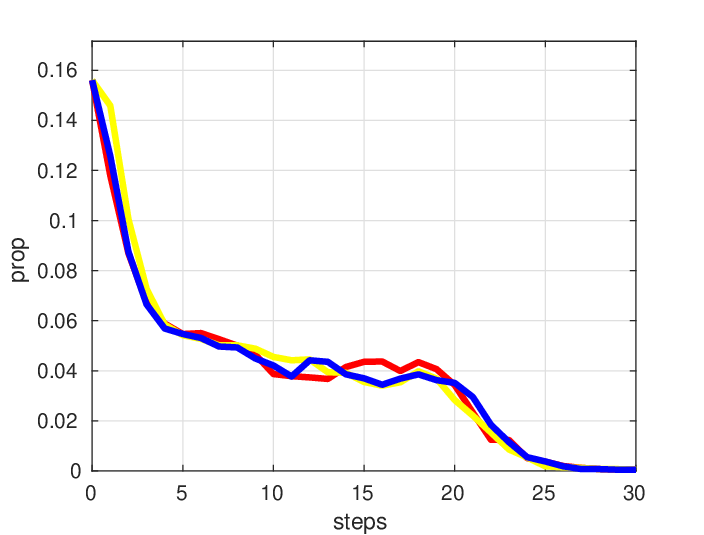}\hspace*{5mm}
  \includegraphics[width=\toto]{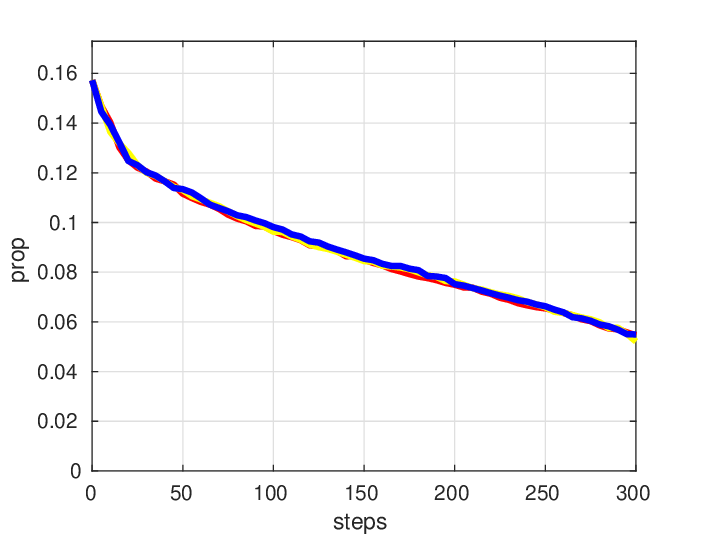}\\
  \includegraphics[width=\toto]{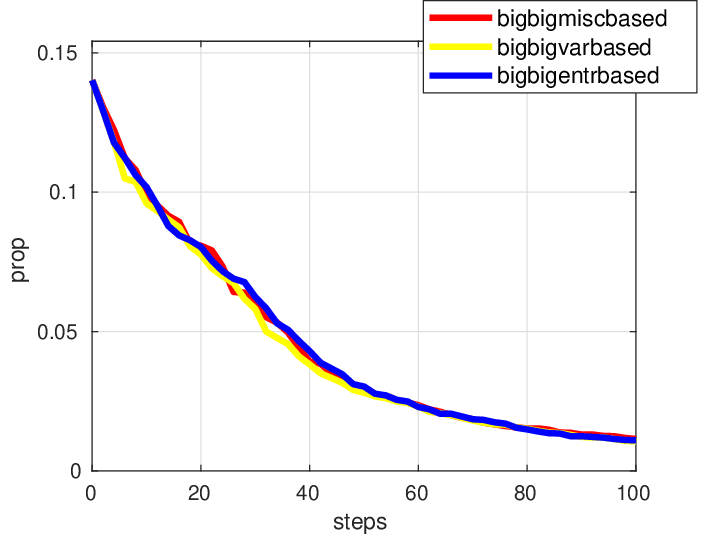}\hspace*{5mm}
  \includegraphics[width=\toto]{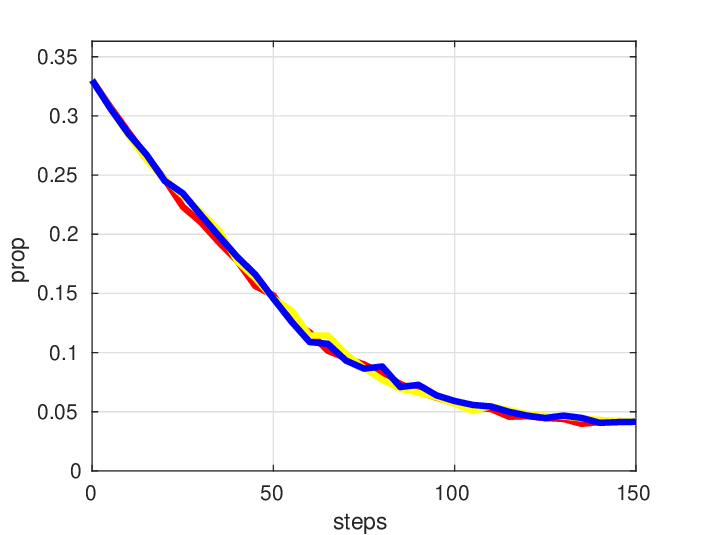}

  \caption{%
    Median of the proportion of misclassified points vs.\ number of
    steps, for 100~repetitions of the algorithms on the test
    functions~$f_1$ (top left), $f_2$ (top right), $f_3$ (bottom left)
    and the volcano case (bottom right).}
  \label{SM:fig:results_comp}
\end{figure}

\section{Complementary results for the examples in the article}
\label{SM:sec:results-experiences}

In this section, some complementary details on the numerical
experiments of \autoref{sec:numerical} are given. %
This include, for all the competitors, the 100~sample paths and the
quantiles of order~$75\%$ and~$95\%$ of the error (proportion of
misclassified points) as a function of the number of steps.

\subsection{\texorpdfstring{Synthetic example $f_1$}{Synthetic example f1}}

\begin{figure}[p]
  \includegraphics[width=\toto]{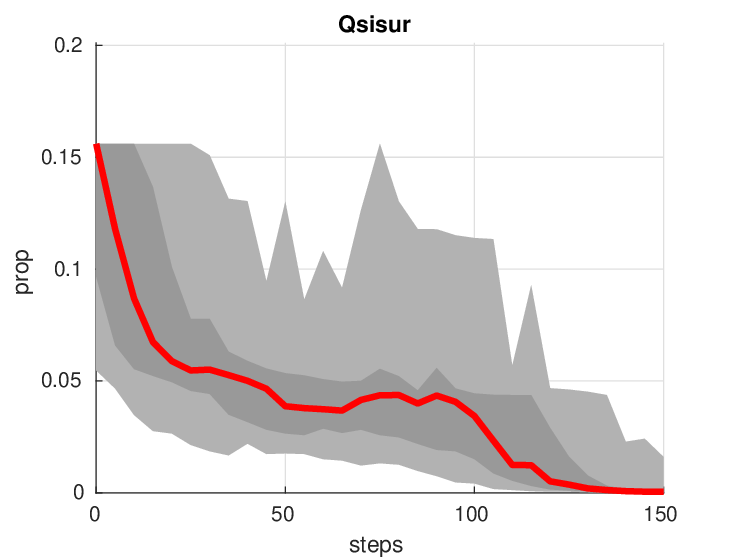}\hspace*{5mm}
  \includegraphics[width=\toto]{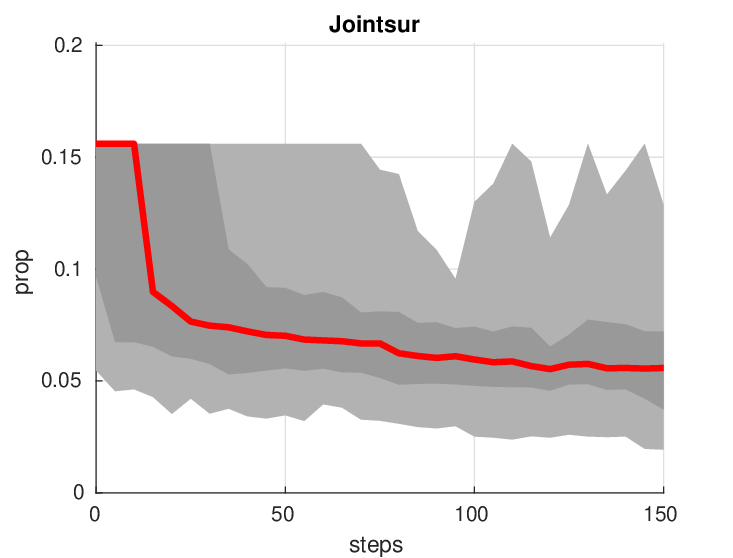}\\[5mm]
  \includegraphics[width=\toto]{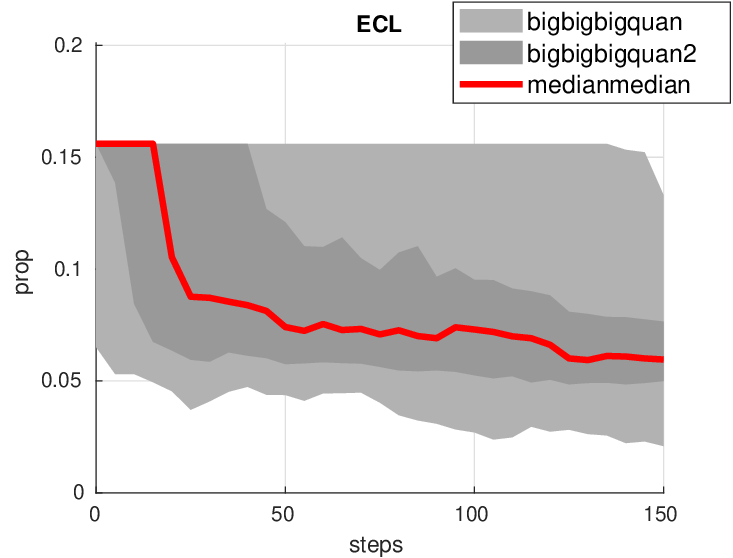}\hspace*{5mm}
  \includegraphics[width=\toto]{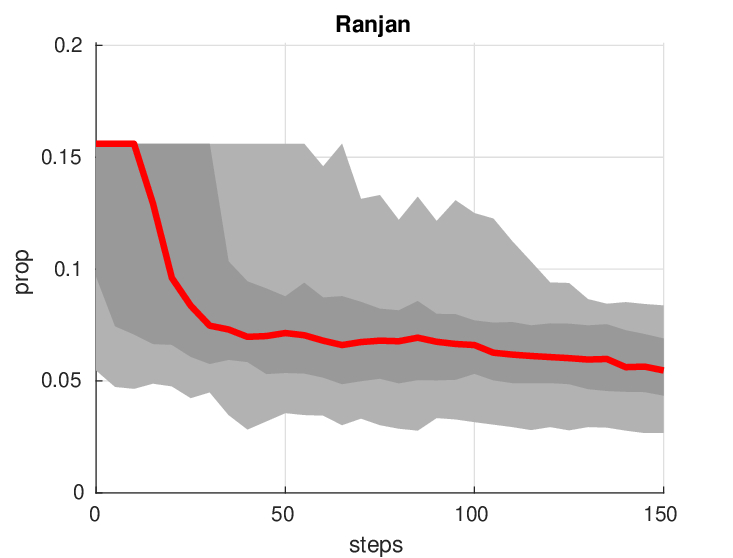}\\[5mm]
  \includegraphics[width=\toto]{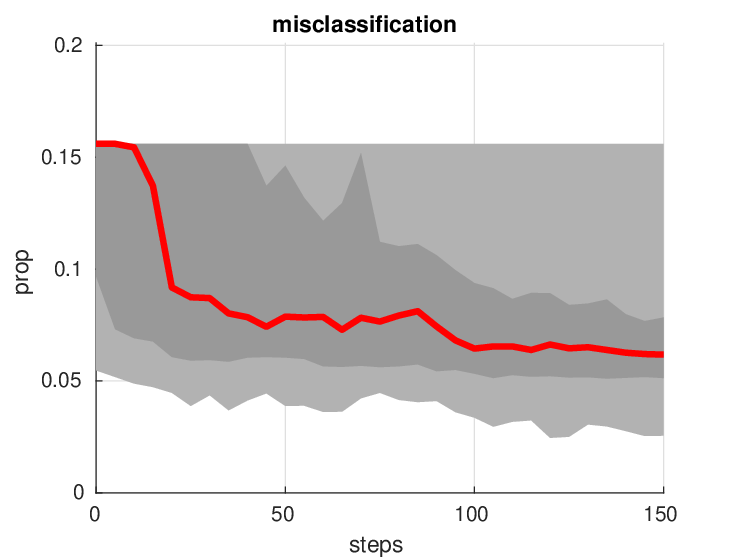}\hspace*{5mm}
  \includegraphics[width=\toto]{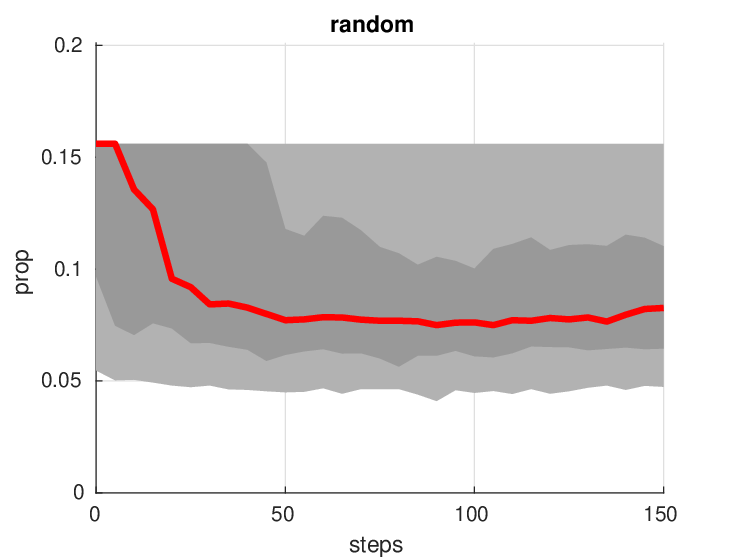}
  \caption{%
    Median and several quantiles of the proportion of misclassified
    points vs.\ number of steps, for 100~repetitions of the algorithms
    on the test function~$f_1$.}
  \label{SM:fig:stats_results_f1}
\end{figure}

\begin{figure}[p]
  \includegraphics[width=\toto]{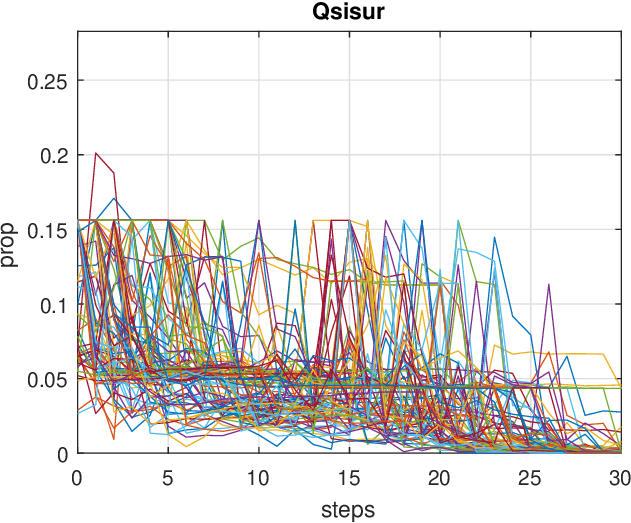}\hspace*{5mm}
  \includegraphics[width=\toto]{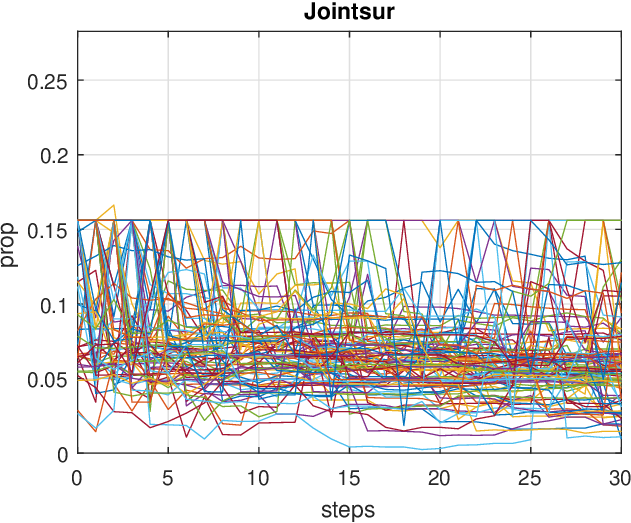}\\[5mm]
  \includegraphics[width=\toto]{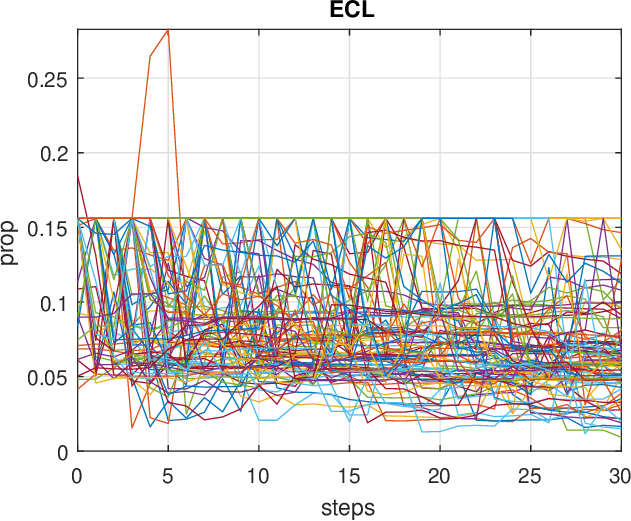}\hspace*{5mm}
  \includegraphics[width=\toto]{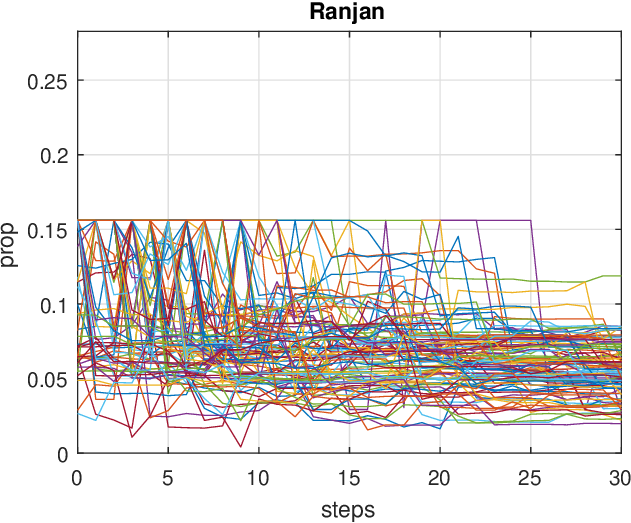}\\[5mm]
  \includegraphics[width=\toto]{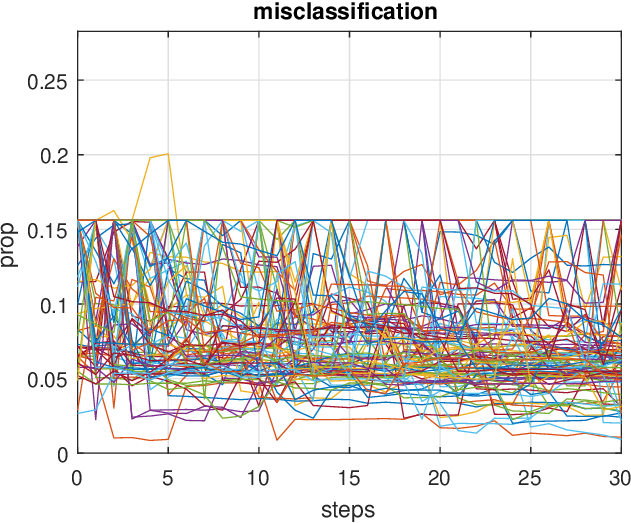}\hspace*{5mm}
  \includegraphics[width=\toto]{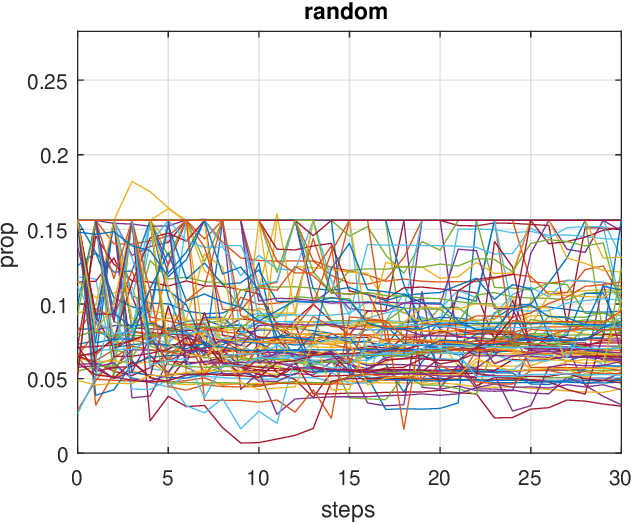}
  \caption{%
    Different sample paths of the proportion of misclassified points
    vs.\ number of steps, for 100~repetitions of the algorithms on the
    test function~$f_1$.}
  \label{SM:fig:trajs_results_f1}
\end{figure}

See Figures~\ref{SM:fig:stats_results_f1_2}--\ref{SM:fig:trajs_results_f1}.

\begin{figure}[p]
  \includegraphics[width=\toto]{graphs/metric_75_branin_mod.eps}\hspace*{5mm}
  \includegraphics[width=\toto]{graphs/metric_95_branin_mod.eps}
  \caption{%
    Quantiles of level~$0.75$ and~$0.95$ for the proportion of
    misclassified points vs.\ number of steps, for 100~repetitions of
    the algorithms on the test function~$f_1$.}
  \label{SM:fig:stats_results_f1_2}
\end{figure}

\subsection{\texorpdfstring{Synthetic example $f_2$}{Synthetic example f2}}

\begin{figure}[p]

  \psfrag{0}     {}
  \psfrag{0.05}  {}
  \psfrag{50}    {}
  \psfrag{100}   {}

  \includegraphics[width=\toto]{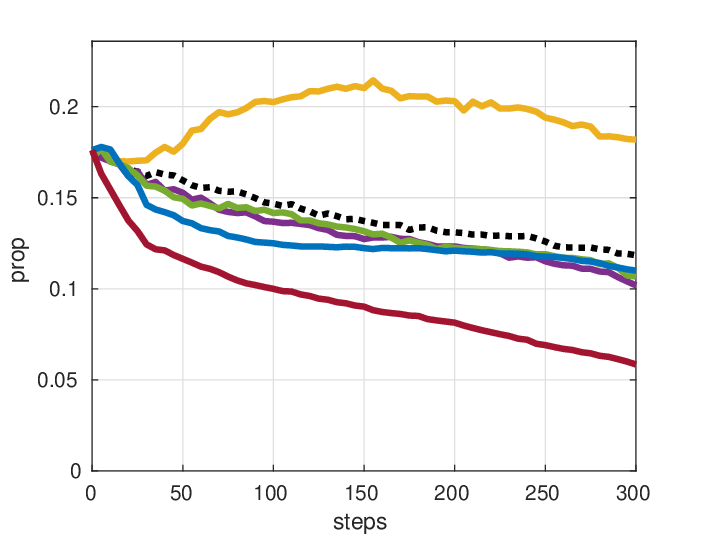}\hspace*{5mm}
  \includegraphics[width=\toto]{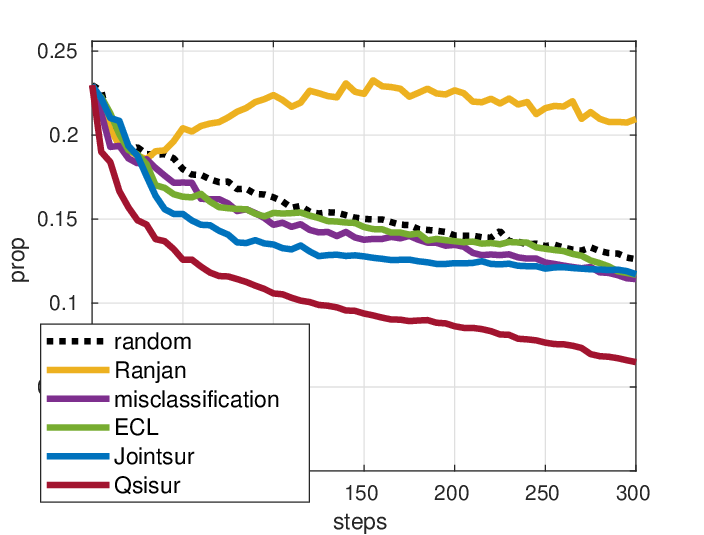}
  \caption{%
    Quantiles of level~$0.75$ and~$0.95$ for the proportion of
    misclassified points vs.\ number of steps, for 100~repetitions of
    the algorithms on the test function~$f_2$.}
  \label{SM:fig:stats_results_f2_2}
\end{figure}

\begin{figure}[p]
  \includegraphics[width=\toto]{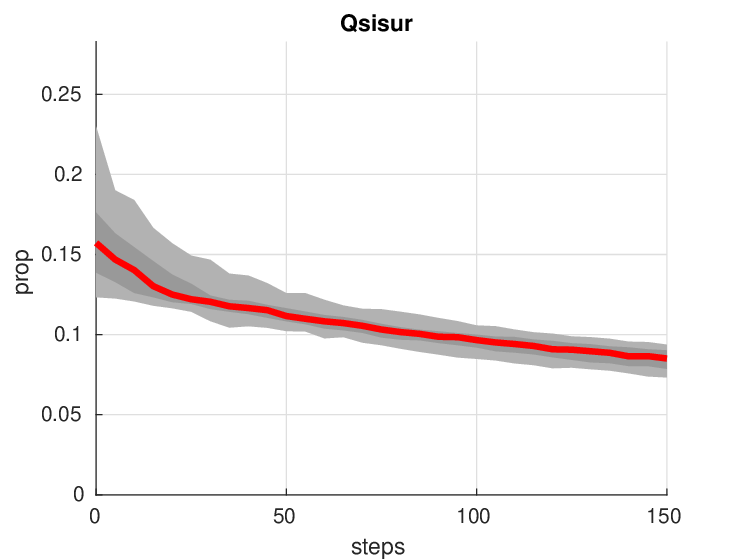}\hspace*{5mm}
  \includegraphics[width=\toto]{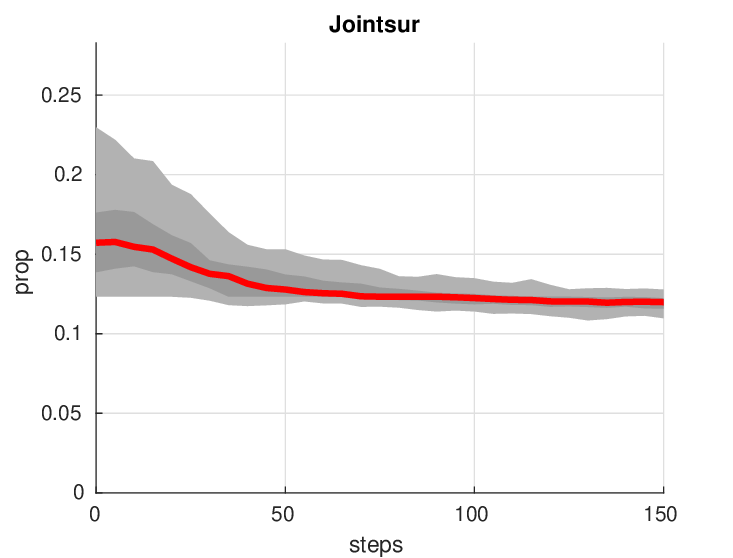}\\[5mm]
  \includegraphics[width=\toto]{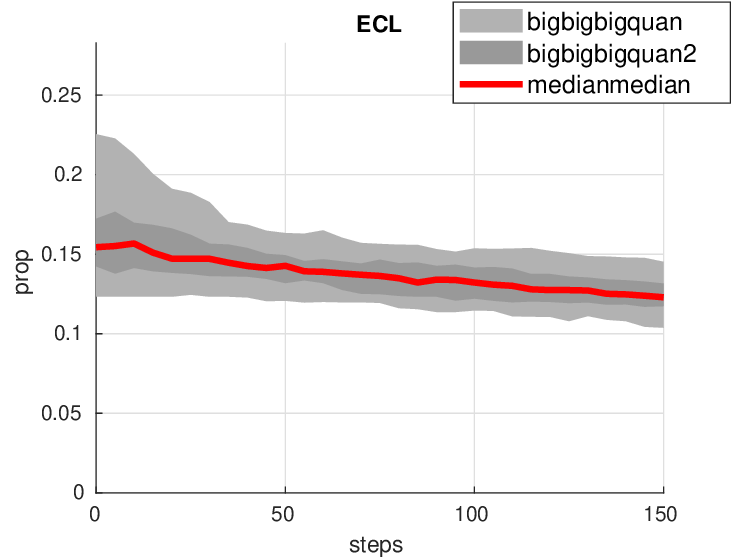}\hspace*{5mm}
  \includegraphics[width=\toto]{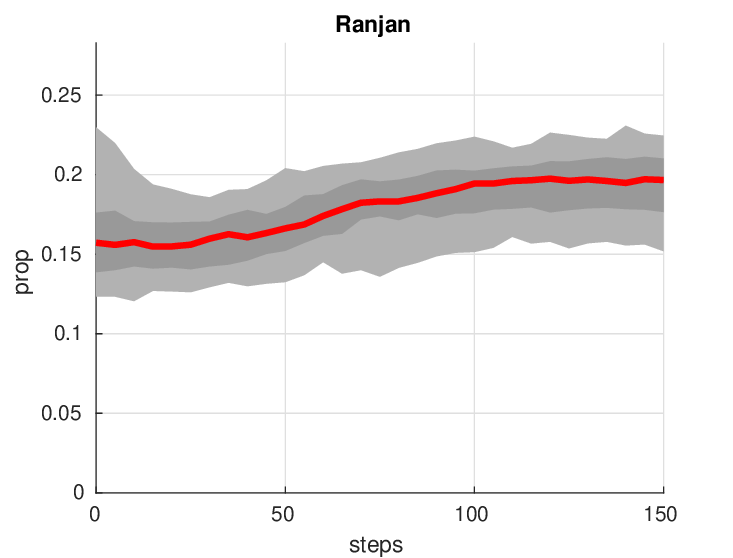}\\[5mm]
  \includegraphics[width=\toto]{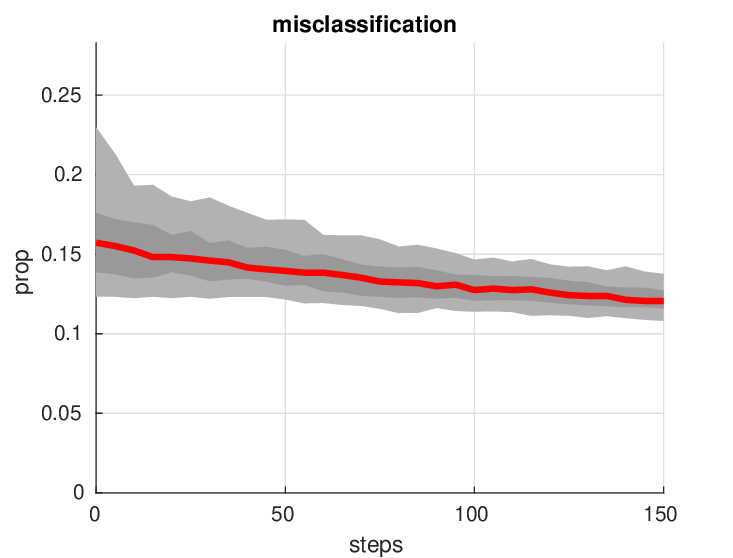}\hspace*{5mm}
  \includegraphics[width=\toto]{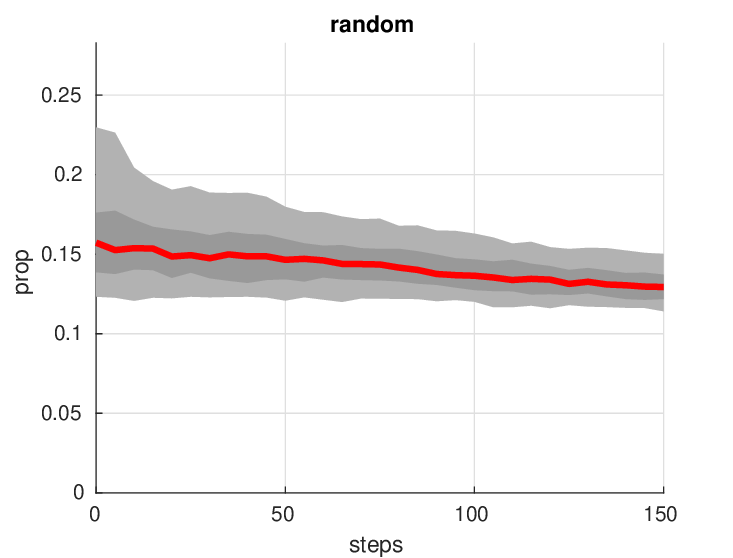}
  \caption{%
    Median and several quantiles of the proportion of misclassified
    points vs.\ number of steps, for 100~repetitions of the algorithms
    on the test function~$f_2$.}
  \label{SM:fig:stats_results_f2}
\end{figure}

\begin{figure}[p]
  \includegraphics[width=\toto]{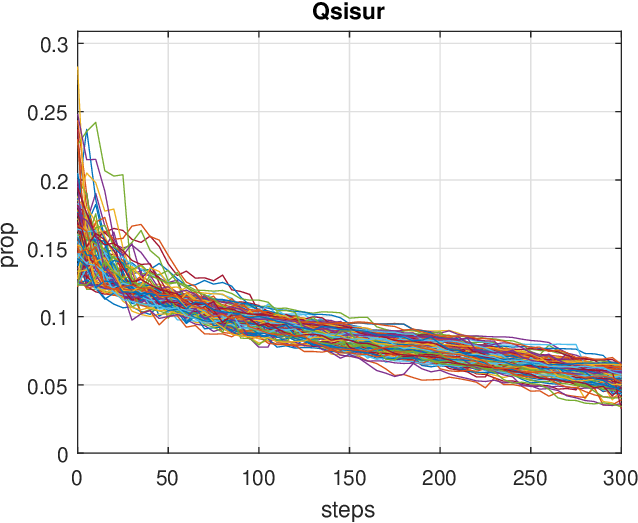}\hspace*{5mm}
  \includegraphics[width=\toto]{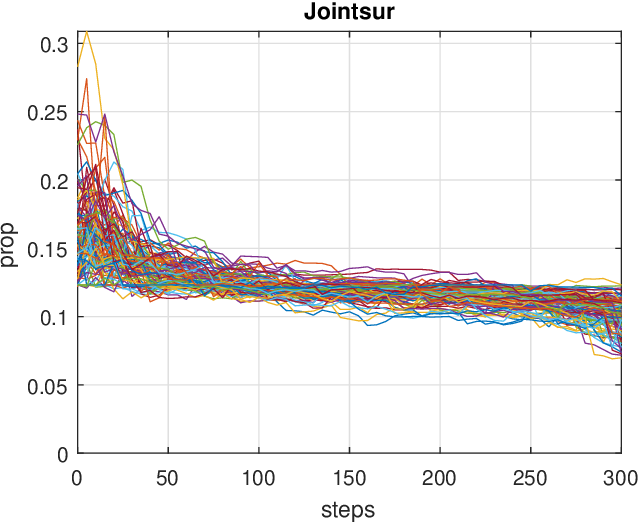}\\[5mm]
  \includegraphics[width=\toto]{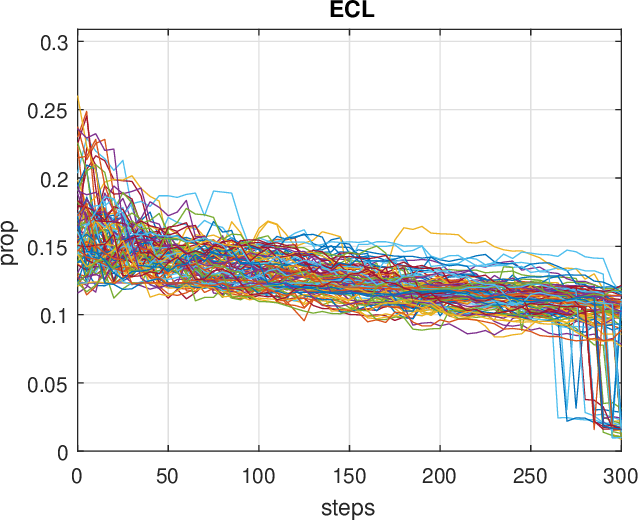}\hspace*{5mm}
  \includegraphics[width=\toto]{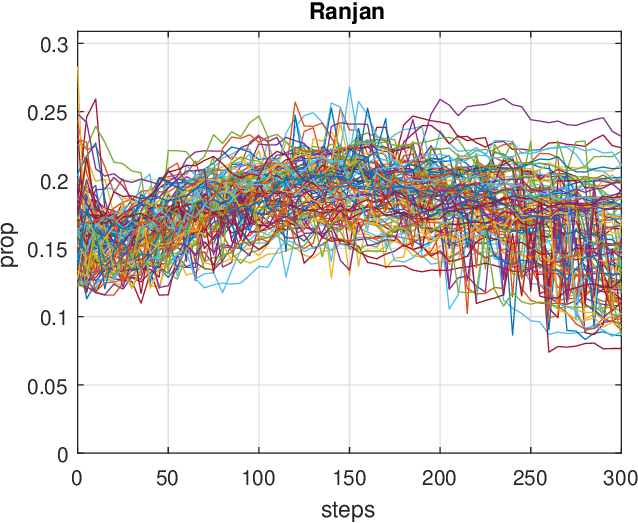}\\[5mm]
  \includegraphics[width=\toto]{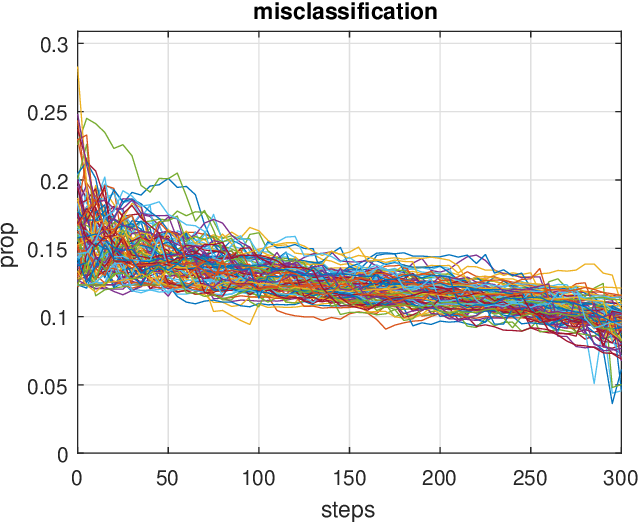}\hspace*{5mm}
  \includegraphics[width=\toto]{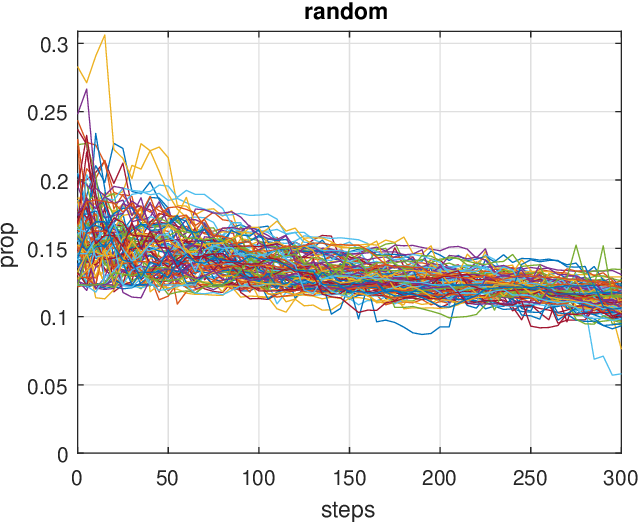}
  \caption{%
    Different sample paths of the proportion of misclassified points
    vs.\ number of steps, for 100~repetitions of the algorithms on the
    test function~$f_2$.}
  \label{SM:fig:trajs_results_f2}
\end{figure}

See Figures~\ref{SM:fig:stats_results_f2_2}--\ref{SM:fig:trajs_results_f2}.

\subsection{\texorpdfstring{Synthetic example $f_3$}{Synthetic example f3}}

\begin{figure}[p]
  \includegraphics[width=\toto]{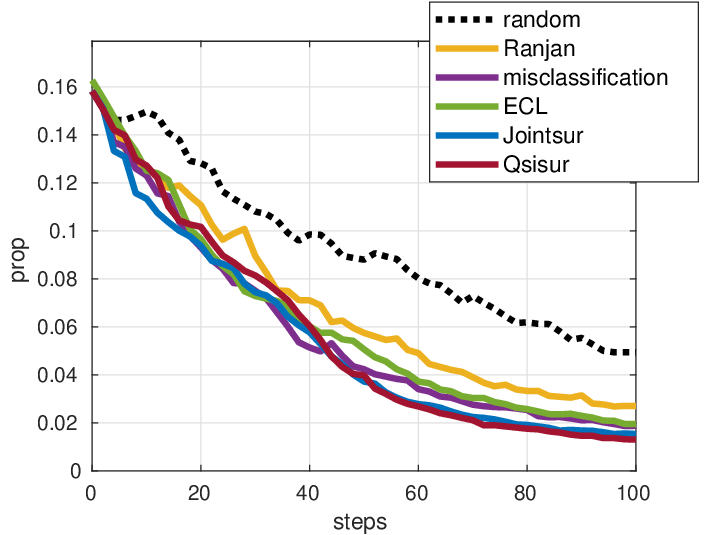}\hspace*{5mm}
  \includegraphics[width=\toto]{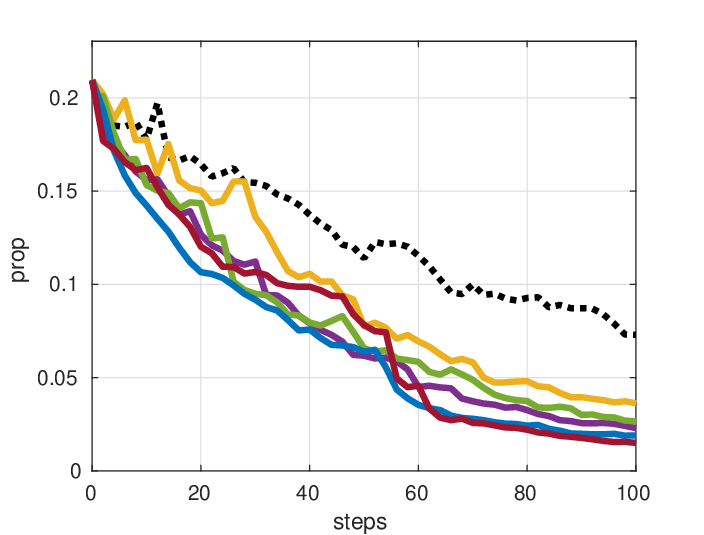}
  \caption{%
    Quantiles of level~$0.75$ and~$0.95$ for the proportion of
    misclassified points vs.\ number of steps, for 100~repetitions of
    the algorithms on the test function~$f_3$.}
  \label{SM:fig:stats_results_f3_2}
\end{figure}

\begin{figure}[p]
  \includegraphics[width=\toto]{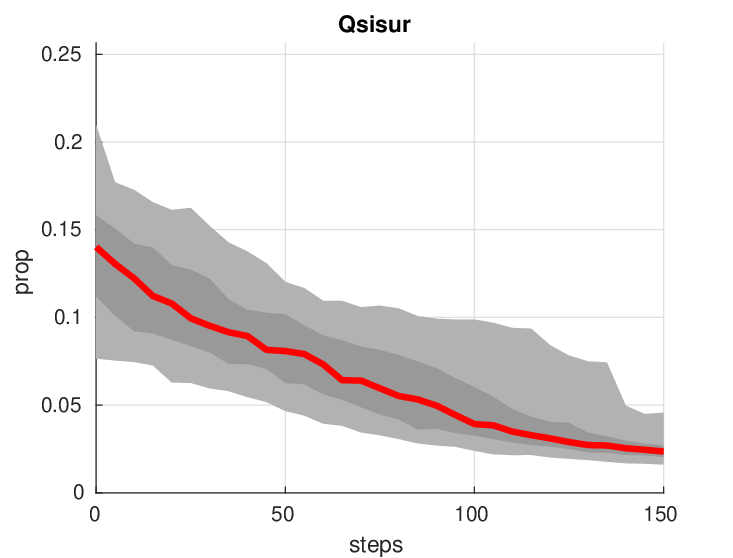}\hspace*{5mm}
  \includegraphics[width=\toto]{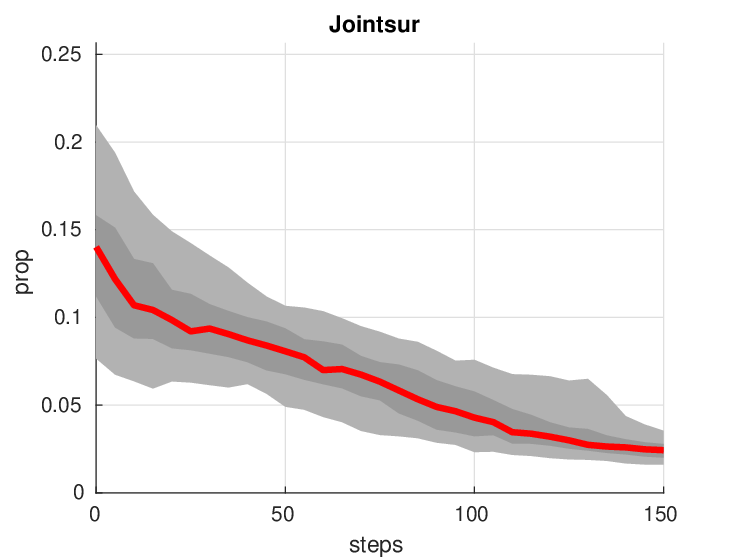}\\[5mm]
  \includegraphics[width=\toto]{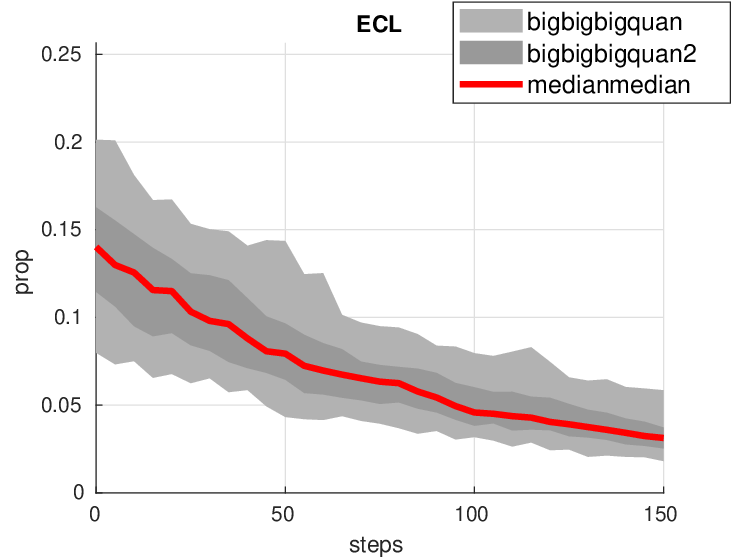}\hspace*{5mm}
  \includegraphics[width=\toto]{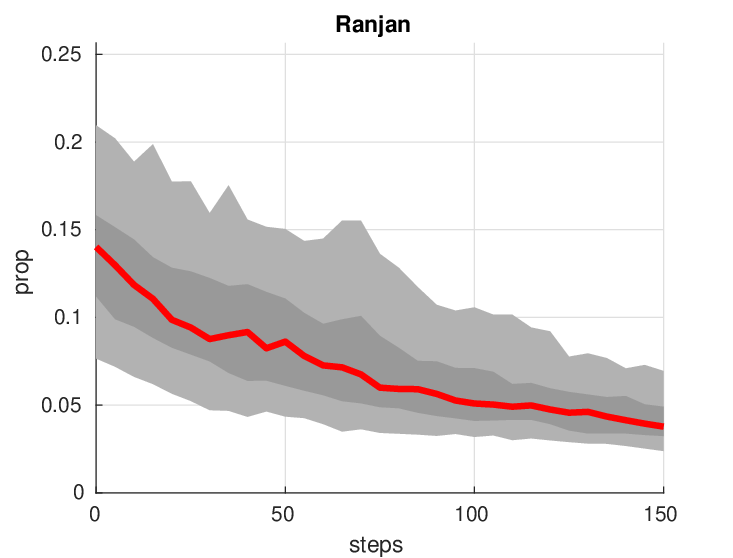}\\[5mm]
  \includegraphics[width=\toto]{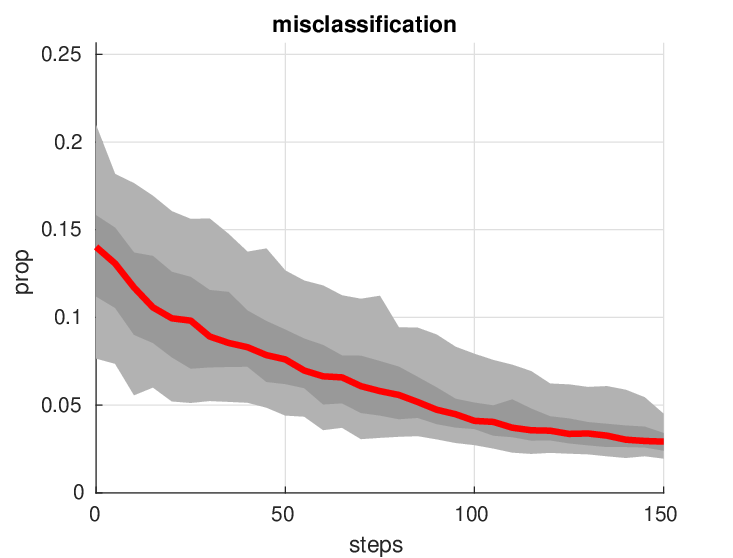}\hspace*{5mm}
  \includegraphics[width=\toto]{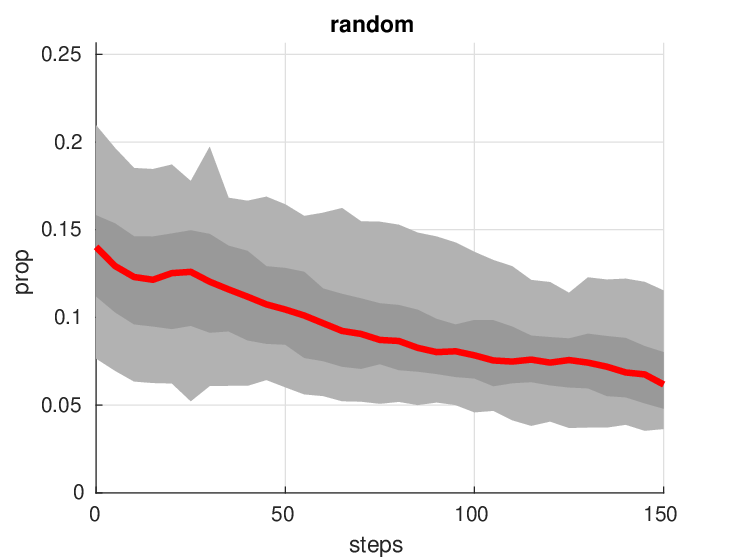}
  \caption{%
    Median and several quantiles of the proportion of misclassified
    points vs.\ number of steps, for 100~repetitions of the algorithms
    on the test function~$f_3$.}
  \label{SM:fig:stats_results_f3}
\end{figure}

\begin{figure}[p]
  \includegraphics[width=\toto]{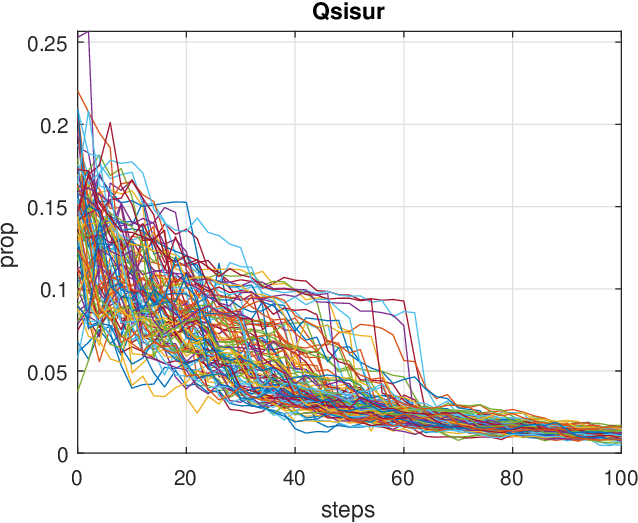}\hspace*{5mm}
  \includegraphics[width=\toto]{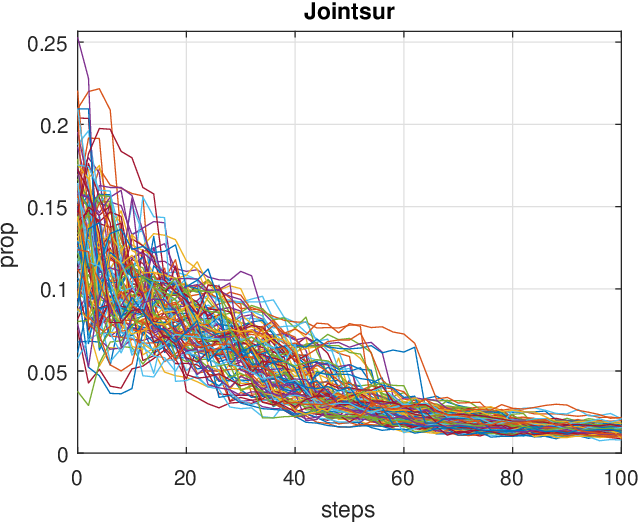}\\[5mm]
  \includegraphics[width=\toto]{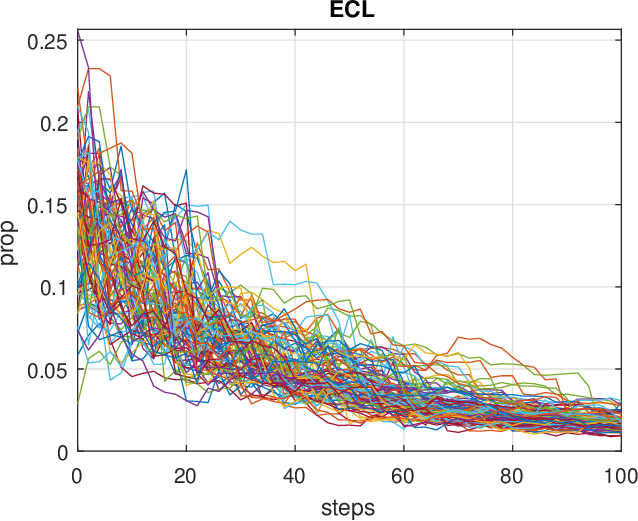}\hspace*{5mm}
  \includegraphics[width=\toto]{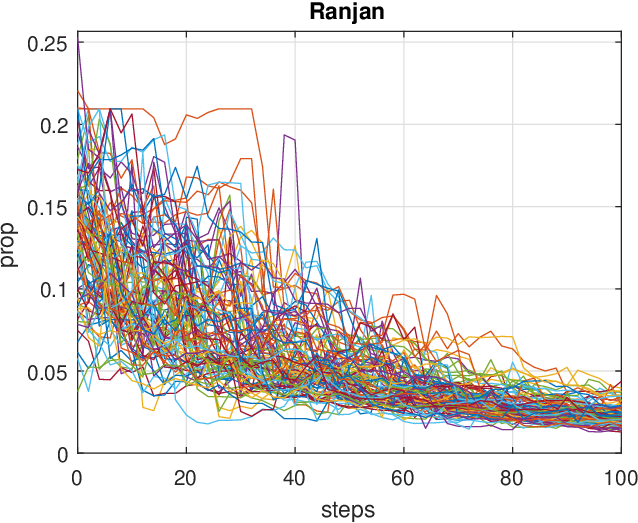}\\[5mm]
  \includegraphics[width=\toto]{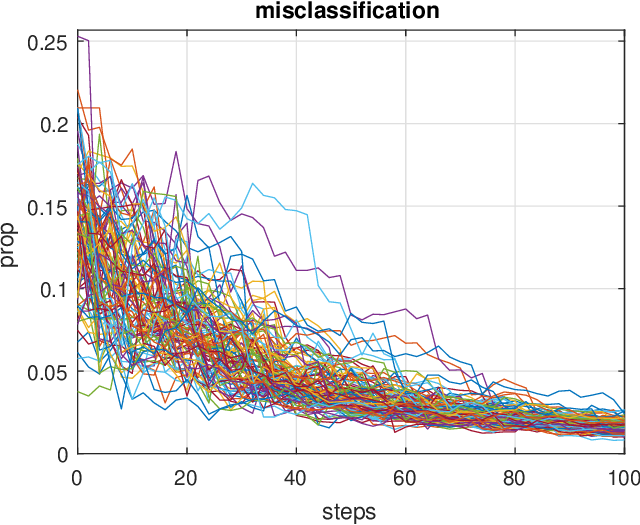}\hspace*{5mm}
  \includegraphics[width=\toto]{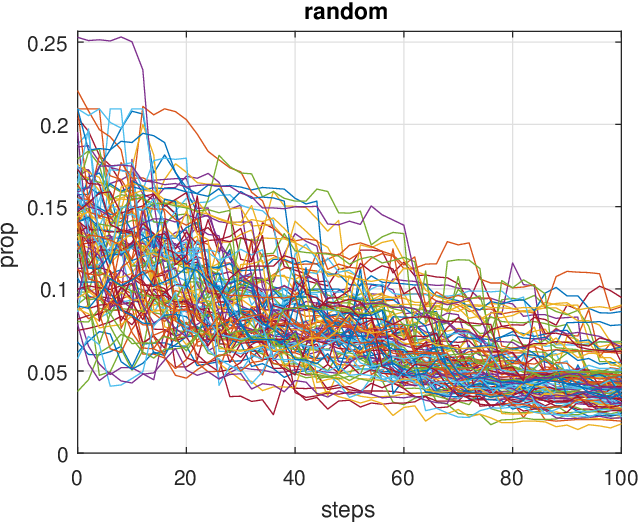}
  \caption{%
    Different sample paths of the proportion of misclassified points
    vs.\ number of steps, for 100~repetitions of the algorithms on the
    test function~$f_3$.}
  \label{SM:fig:trajs_results_f3}
\end{figure}

See Figures~\ref{SM:fig:stats_results_f3_2}--\ref{SM:fig:trajs_results_f3}.

\subsection{Volcano test case}

\begin{figure}[p]
  \includegraphics[width=\toto]{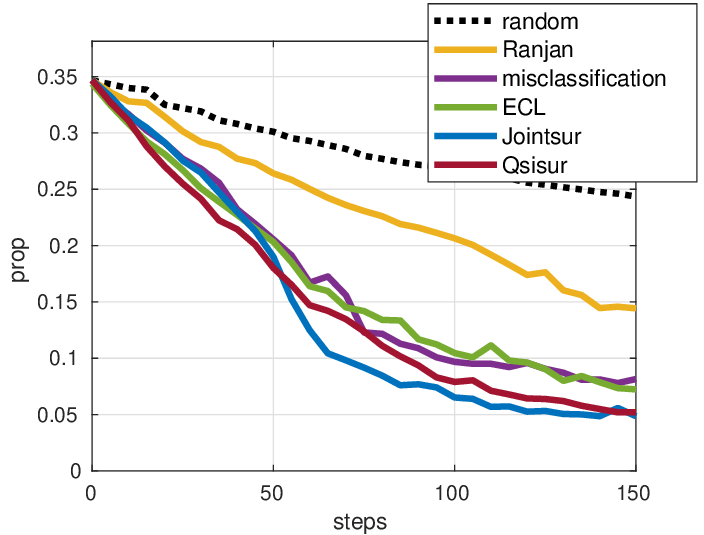}\hspace*{5mm}
  \includegraphics[width=\toto]{graphs/metric_95_volcano.eps}
  \caption{%
    Quantiles of level~$0.75$ and~$0.95$ for the proportion of
    misclassified points vs.\ number of steps, for 100~repetitions of
    the algorithms on the test case volcano.}
  \label{SM:fig:stats_results_volc_2}
\end{figure}

\begin{figure}[p]
  \includegraphics[width=\toto]{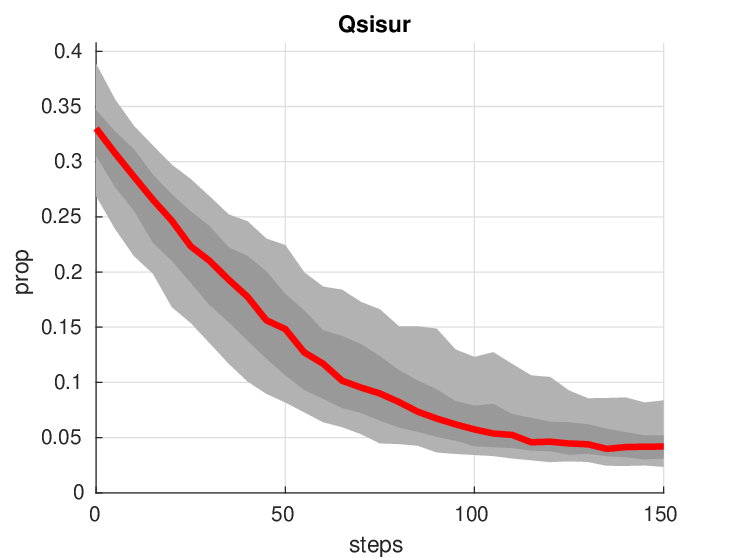}\hspace*{5mm}
  \includegraphics[width=\toto]{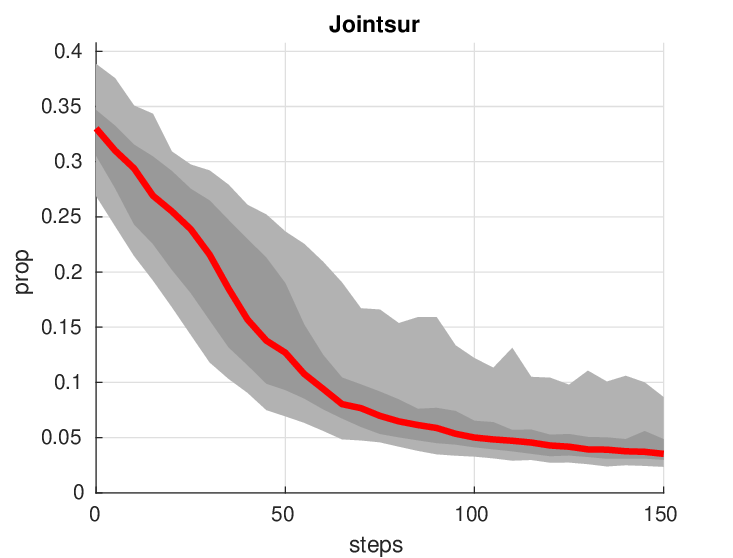}\\[5mm]
  \includegraphics[width=\toto]{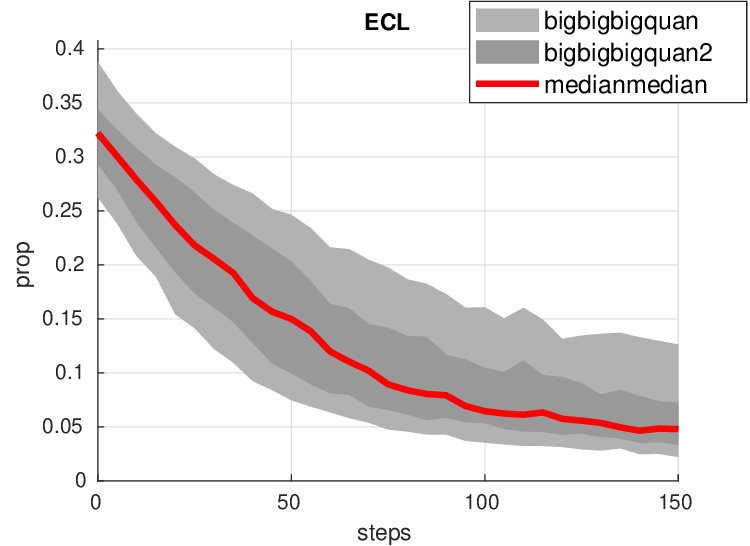}\hspace*{5mm}
  \includegraphics[width=\toto]{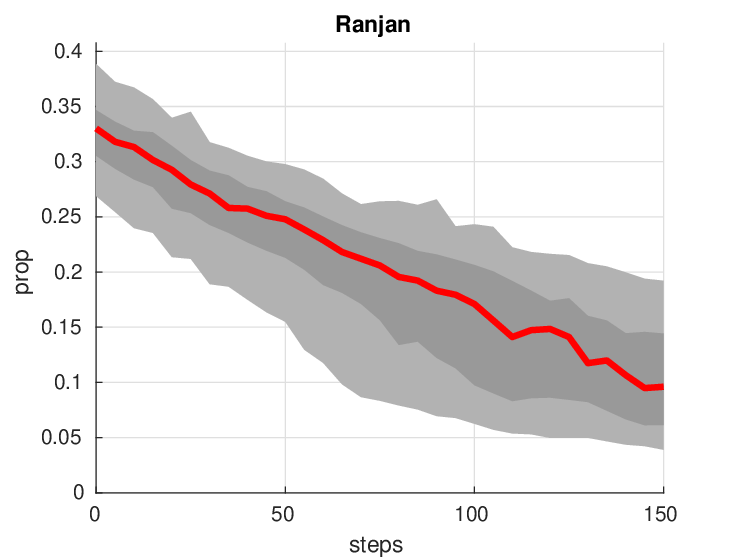}\\[5mm]
  \includegraphics[width=\toto]{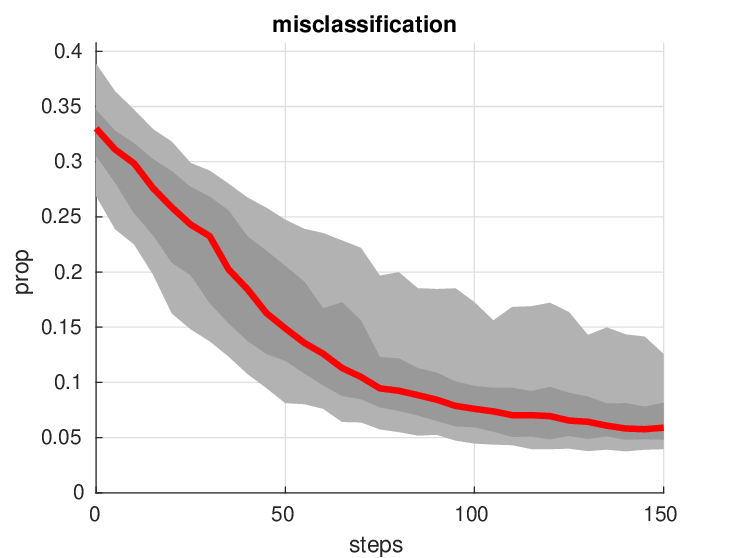}\hspace*{5mm}
  \includegraphics[width=\toto]{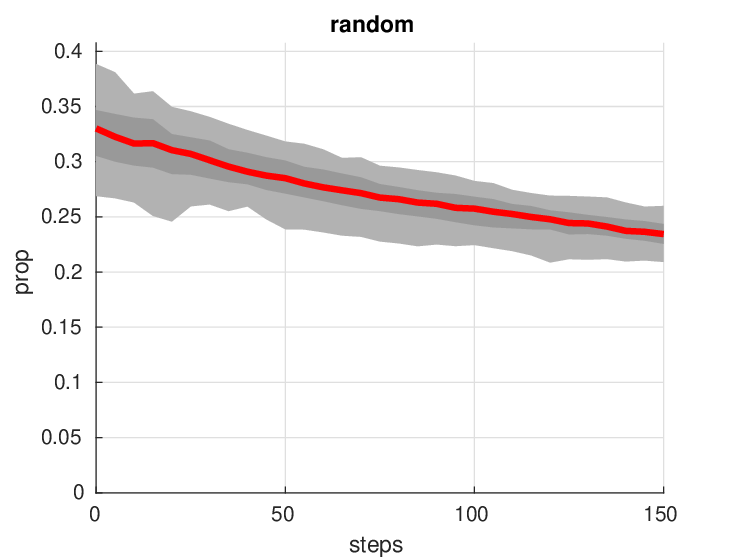}
  \caption{%
    Median and several quantiles of the proportion of misclassified
    points vs.\ number of steps, for 100~repetitions of the algorithms
    on the test case volcano.}
  \label{SM:fig:stats_results_volc}
\end{figure}

\begin{figure}[p]
  \includegraphics[width=\toto]{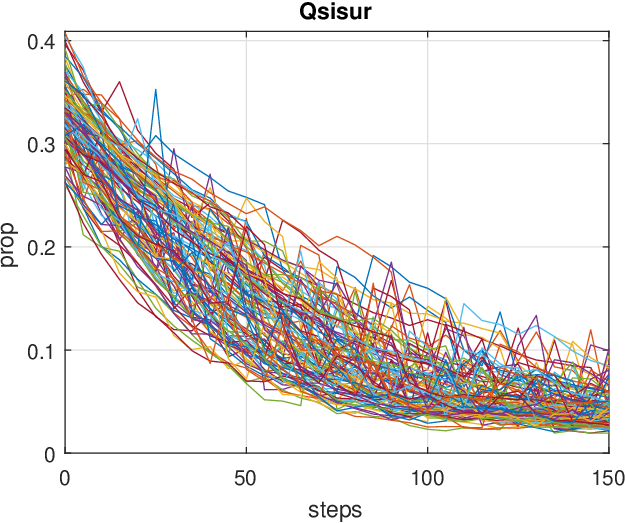}\hspace*{5mm}
  \includegraphics[width=\toto]{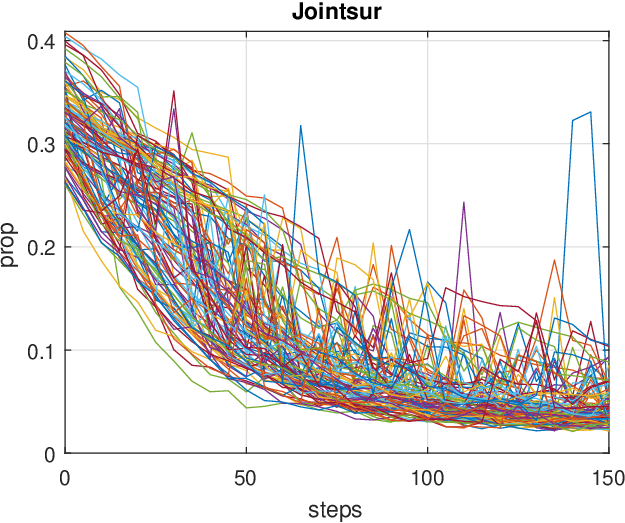}\\[5mm]
  \includegraphics[width=\toto]{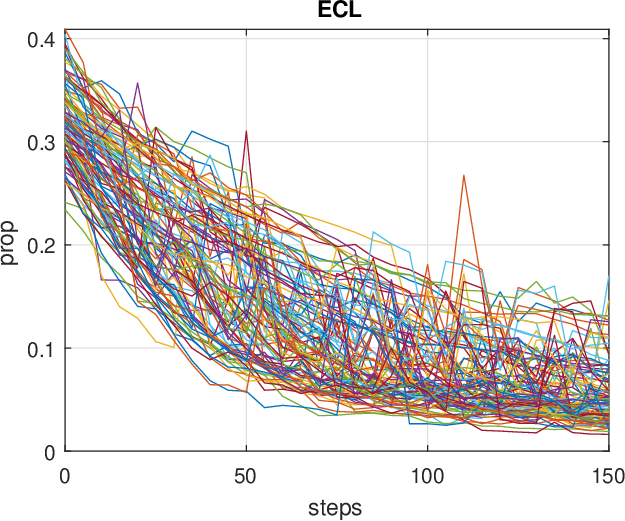}\hspace*{5mm}
  \includegraphics[width=\toto]{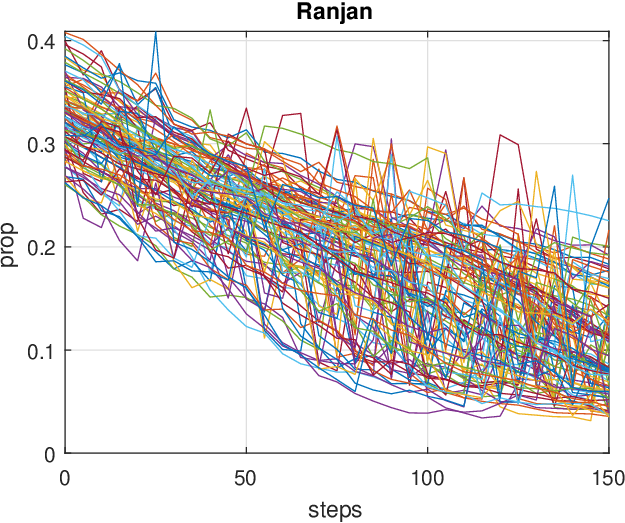}\\[5mm]
  \includegraphics[width=\toto]{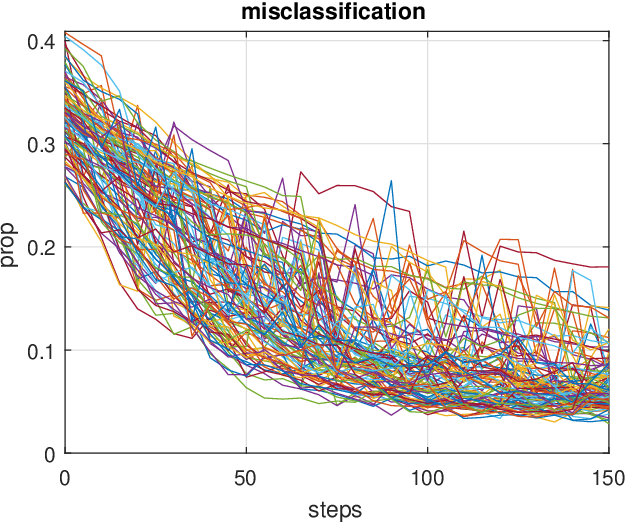}\hspace*{5mm}
  \includegraphics[width=\toto]{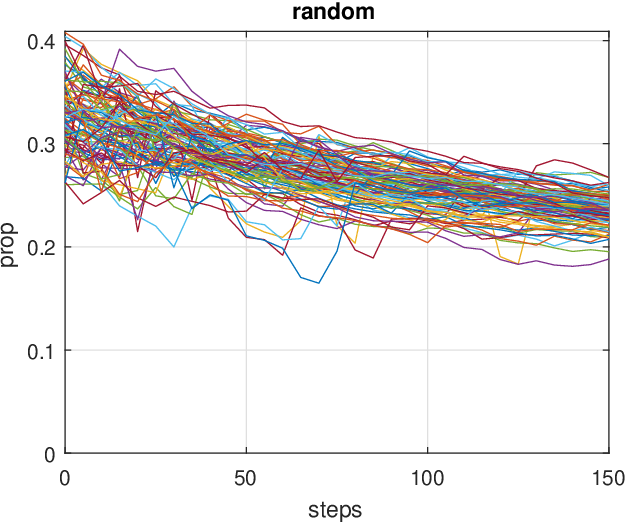}
  \caption{%
    Different sample paths of the proportion of misclassified points
    vs. number of steps, for 100~repetitions of the algorithms on the
    test case volcano.}
  \label{SM:fig:trajs_results_volc}
\end{figure}

See Figures~\ref{SM:fig:stats_results_volc_2}--\ref{SM:fig:trajs_results_volc}.

\end{document}